\documentclass{article}
\usepackage[preprint,eandd]{neurips_2026}

\usepackage[greek,russian,spanish,english]{babel}
\usepackage[utf8]{inputenc}
\usepackage[T1]{fontenc}
\usepackage{hyperref}
\hypersetup{colorlinks=true,citecolor=blue,linkcolor=blue,urlcolor=blue}
\usepackage{url}
\usepackage{graphicx}
\usepackage{booktabs}

\usepackage{newfloat}
\usepackage{listings}
\usepackage{amsmath}
\usepackage{amssymb}
\usepackage{amsthm}
\usepackage{amscd}
\usepackage{multirow}
\usepackage{subcaption}
\usepackage{wrapfig}
\usepackage{placeins}
\usepackage{dblfloatfix}
\usepackage{tcolorbox}
\usepackage{enumitem}
\usepackage{xcolor}
\usepackage{newunicodechar}

\newunicodechar{’}{'}
\newunicodechar{‘}{`}
\newunicodechar{“}{``}
\newunicodechar{”}{''}
\newunicodechar{—}{---}
\newunicodechar{–}{--}
\newunicodechar{°}{$^\circ$}

\usepackage{etoolbox}
\makeatletter
\patchcmd{\@toptitlebar}{\vskip 0.25in}{\vskip 0.14in}{}{}
\patchcmd{\@bottomtitlebar}{\vskip 0.29in}{\vskip 0.16in}{}{}
\patchcmd{\@maketitle}{\vskip 0.3in}{\vskip 0.10in}{}{}
\makeatother

\AtBeginDocument{%
  }

\begin{document}

\title{LG-Bench: A Graph-Structured Evaluation Benchmark for Life Sciences}

\author{
  \textbf{Lu Sun\textsuperscript{1}}\thanks{Equal contribution.},
  \textbf{Xiangyi Zhang\textsuperscript{4}}\footnotemark[1],
  \textbf{Xiangyang Zhu\textsuperscript{1}},
  \textbf{Zijian Chen\textsuperscript{1,2}},
  \\
  \textbf{Qiang Hu\textsuperscript{3}},
  \textbf{Yuan Tian\textsuperscript{1}}\thanks{Corresponding author.},
  \textbf{Guangtao Zhai\textsuperscript{1,2}}
  \\
  \textsuperscript{1}Shanghai Artificial Intelligence Laboratory \\
  \textsuperscript{2}Institute of Image Communication and Network Engineering, Shanghai Jiao Tong University\\
  \textsuperscript{3}Cooperative Medianet Innovation Center, Shanghai Jiao Tong University\\
  \textsuperscript{4}SJTU-Paris Elite Institute of Technology, Shanghai Jiao Tong University\\
  \strut\\
  \small{\textbf{Correspondence:} \href{mailto:tianyuan168326@outlook.com}{tianyuan168326@outlook.com}}
}

\maketitle

\begin{abstract}
    Traditional evaluation benchmarks reduce inherently interconnected scientific knowledge in life sciences into flat lists of questions, disregarding the underlying topological structure of the knowledge. We introduce LG-Bench, the first graph-structured benchmark for life sciences, featuring over 10,000 high-quality multiple-choice questions across medicine, biology, and chemistry. Our approach constructs a weighted evaluation graph using bidirectional matching and semantic similarity algorithms, where nodes represent questions and edge weights capture their semantic relationships. Leveraging this graph topology, we design two novel evaluation metrics. The Global Coherence Score (GCS) measures a model’s consistency within semantically related neighborhoods, while Knowledge Balance Score (KBS) analyzes how model errors are distributed across the graph to reveal conceptual blind spots. LG-Bench facilitates fine-grained comparison of LLMs by surfacing differences in conceptual coherence and patterns of knowledge organization across models. Our framework shifts the evaluation paradigm from flat accuracy metrics to structure-aware analysis, offering a new lens for diagnosing and improving LLM performance in the life sciences domain.
\end{abstract}

\section{Introduction}
\fussy
\sloppy

\begin{wrapfigure}[15]{r}{.5\linewidth}
    \vspace{-\baselineskip}
    \centering
    \captionsetup{font=small}
    \includegraphics[width=0.85\linewidth]{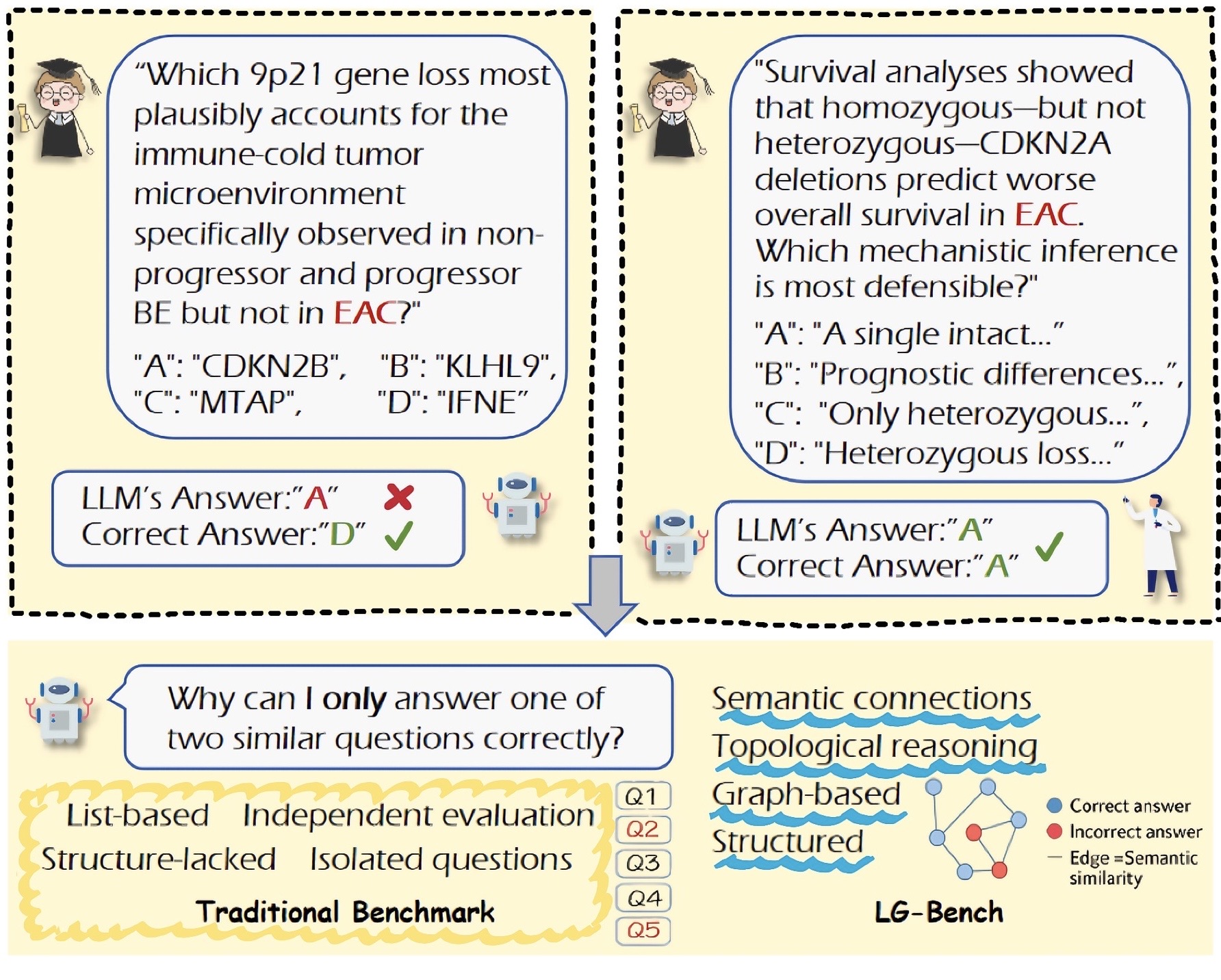}
    \caption{Case study comparing LG-Bench with traditional benchmarks. Unlike traditional list-based approaches that overlook semantic connections, LG-Bench utilizes a graph structure to capture these links, effectively revealing LLM inconsistencies on related questions.}
    \label{fig:0}
\end{wrapfigure}

Recent breakthroughs in large language models (LLMs) epitomized by systems have rapidly accelerated progress in natural language processing and sparked an intense wave of research activity across multiple scientific disciplines. As model scale and performance continue to increase, the community is increasingly dependent on systematic evaluation to guide development, compare approaches, and guarantee safe deployment.

\begin{figure}[t]
    \centering
    \includegraphics[width=0.92\linewidth]{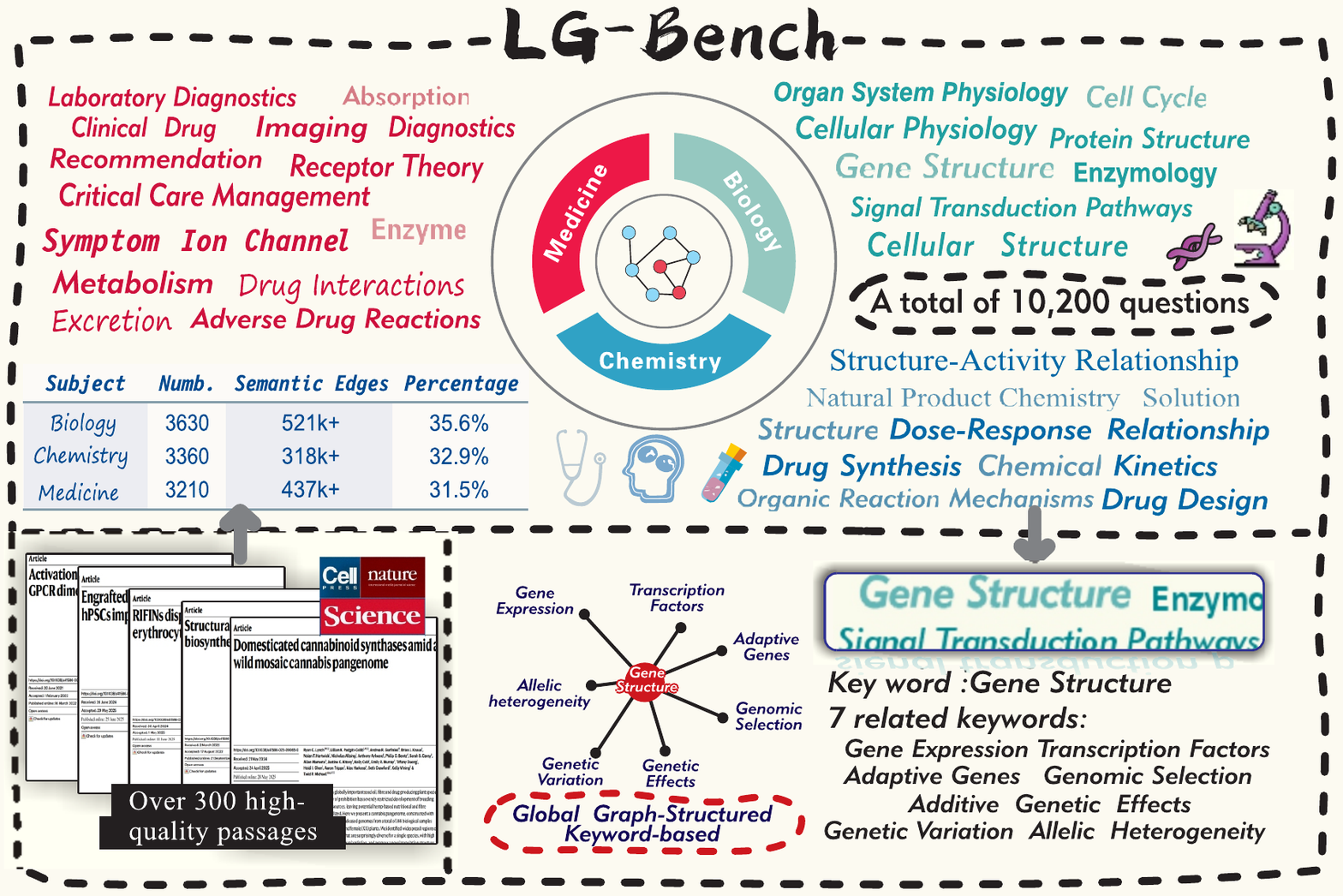}
    \caption{Overview diagram of LG-Bench. Our LG-Bench extracts over 10,000 questions from top-tier journals in the life sciences, covering three subfields: Medicine, Biology, and Chemistry. It innovatively introduces graph structures into the benchmark to better capture the nuanced capabilities of large language models.}
    \label{fig:1}
\end{figure}

However, current evaluations such as practices in artificial intelligence suffer from a fundamental misrepresentation: they treat knowledge as isolated, independent facts when it naturally forms interconnected webs of understanding. This "flat-world illusion" is particularly problematic in complex domains such as life sciences \citep{bodenreider2004unified,jin2021disease}, where understanding protein function requires grasping its role in cellular pathways, regulatory mechanisms, and therapeutic implications as an integrated whole. Traditional benchmarks present evaluation questions as flat lists, obscuring the rich semantic relationships that define genuine domain expertise. These limitations make it hard for current metrics to distinguish between genuine understanding and mere pattern matching.

To address this gap, we introduce LG-Bench, the first graph-structured benchmark for life sciences evaluation. As shown in Fig.~\ref{fig:0}, LG-Bench contains over 10k expert-reviewed questions drawn from recent peer-reviewed literature across Medicine, Biology, and Chemistry, with representative sources including Nature, Science, and Cell. The benchmark is designed to cover both foundational knowledge and recent scientific findings, making it suitable for evaluating whether models can organize concepts rather than only recall isolated facts. Using bidirectional matching and semantic similarity, we transform the question corpus into a weighted evaluation graph whose edges capture relationships between semantically related questions. This graph makes local neighborhoods of related scientific concepts explicit, enabling evaluation beyond flat accuracy by measuring knowledge coherence and the topological distribution of model errors.

Our contributions are threefold:

1. We introduce LG-Bench\footnote{An anonymized repository containing the benchmark data, graph file, evaluation code, and Croissant metadata is available at \url{https://anonymous.4open.science/r/LG-Bench-B6D2/}.}, the first graph-structured benchmark for life sciences, containing over 10k expert-curated questions from recent literature from leading scientific journals spanning Medicine, Biology, and Chemistry, as shown in Fig.~\ref{fig:1}.

2. We propose novel graph-based evaluation methods including the Global Coherence Score (GCS) for overall knowledge coherence, and the Knowledge Balance Score (KBS) for variance in local coherence across the graph.

3. We conducted a systematic evaluation of LG-Bench across large language models of varying scales, and leveraged GCS and KBS to deeply analyze performance flaws, providing practical support for capability diagnosis and optimization of large models in the life science.

\section{Related Work}
Traditional evaluation of large language models relies on accuracy-based metrics, treating each question as an independent unit \citep{li2024llms,richard2015burden}. This approach assumes knowledge can be divided into isolated facts, missing the interconnected nature of true expertise. Although recent work explores more nuanced frameworks \citep{mondorf2024beyond,xu2025large,aibench}, current metrics still struggle to differentiate genuine understanding from sophisticated pattern matching, a limitation particularly evident in complex scientific domains.

This challenge is amplified in the life sciences. Prominent benchmarks like UMLS \citep{bodenreider2004unified}, MedQA \citep{jin2021disease}, PubMedQA \citep{jin2019pubmedqa}, MLEC-QA \citep{li2021mlec}, and BioASQ \citep{nentidis2023overview} suffer from critical limitations. They often rely on static knowledge, failing to capture the rapid evolution of research in areas like personalized medicine and advanced therapeutics \citep{cai2024medbench,chen2025evaluating,zhou2025benchmarking}. Furthermore, their narrow focus on medicine over foundational sciences like biology and chemistry restricts the ability to assess comprehensive, interdisciplinary knowledge, which is essential for reliable evaluation.

A promising direction to address these structural flaws lies in graph-based knowledge representation. While knowledge graphs are extensively studied in NLP \citep{hogan2021knowledge,ji2021survey}, and techniques like graph embeddings \citep{bordes2013translating,wang2017knowledge} and graph neural networks \citep{kipf2016semi,wu2020comprehensive} have proven effective at capturing complex relationships, these insights have not yet been applied to benchmark construction. In life sciences, where knowledge inherently forms interconnected networks from molecular interactions to physiological systems, the absence of a graph-structured evaluation framework represents a significant gap. Our work bridges the gap by introducing the first graph-structured benchmark that explicitly models and leverages these knowledge relationships for a more meaningful and robust evaluation.

\section{Graph-Structured Benchmark Construction}
\begin{wrapfigure}[11]{r}{.50\linewidth}
    \vspace{-\baselineskip}
    \centering
    \includegraphics[width=.98\linewidth]{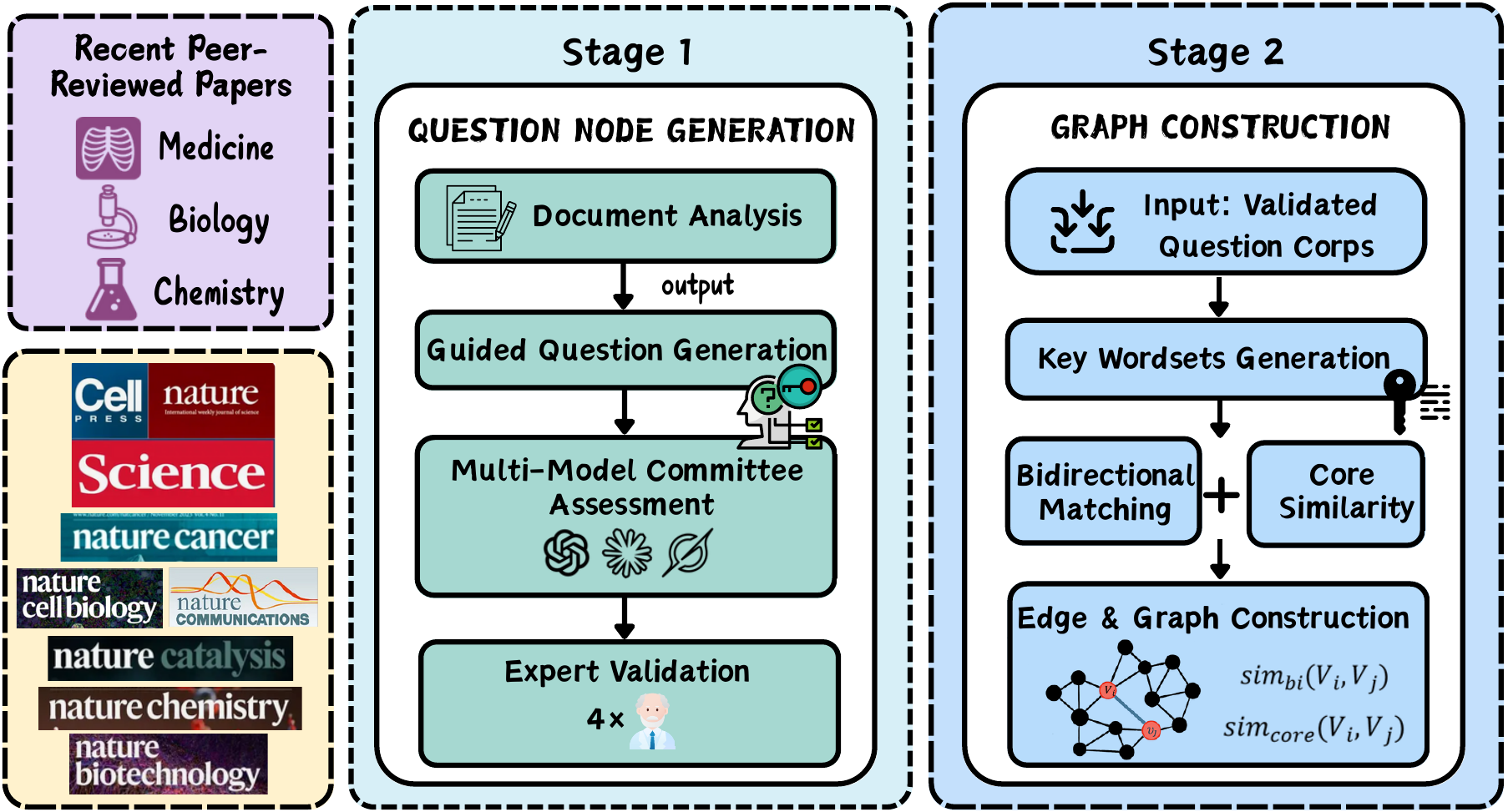}
    \caption{Two-stage Pipeline for Construction of LG-Bench.}
    \label{fig:2}
\end{wrapfigure}

Our benchmark construction follows a two-stage pipeline shown as Fig.~\ref{fig:2}. First, we generate high-quality evaluation questions from recent scientific literature through a sophisticated multi-stage process with expert validation. Second, we construct a weighted evaluation graph, transforming the flat question corpus into a structured representation that reflects the interconnected nature of scientific knowledge.

\subsection{Question Node Generation}
We generate question sets through a sophisticated multi-stage pipeline that transforms cutting-edge scientific literature into high-quality evaluation questions.

\textbf{Document Analysis.} Our pipeline begins with a multi-modal large language model (LLM) analyzer that processes recent peer-reviewed papers from leading journals in the life sciences. This analyzer extracts structured knowledge representations from each document $d \in D_{\text{corpus}}$, identifying core concepts along with their hierarchical and semantic relationships. It systematically extracts knowledge pairs, consisting of concepts and their associated properties or relations, which serve as fundamental units for downstream question generation. The analyzer also assigns each identified concept a corresponding Bloom’s Taxonomy level \citep{forehand2010bloom}, specifying its cognitive depth. The system enforces explicit targets for the distribution of questions across these levels, ensuring a balanced coverage of cognitive complexity in the final benchmark.

\textbf{Guided Question Generation.} The generation stage employs a specialized LLM that receives guidance from the analyzer's output. Each generation prompt includes domain-specific knowledge context and key concepts, a target Bloom level with specific cognitive verbs, scientific accuracy constraints, and requirements for testing conceptual relationships. This guidance ensures generated questions genuinely test integrated understanding rather than isolated fact recall.

\textbf{Multi-Model Committee Assessment.} The quality assurance stage involves a committee of three state-of-the-art large language models (GPT-4o, Claude 3.7 Sonnet, and Grok-3) that independently evaluate each generated question. The committee members assess questions across multiple dimensions, including scientific accuracy, cognitive complexity, and conceptual integration. Through a voting mechanism, questions that fail to meet quality thresholds are either rejected outright due to insufficient quality or routed to mandatory human expert review. Additionally, the committee annotates the relevant knowledge sources and the rationale behind each judgment to facilitate more efficient and accurate expert verification.

\textbf{Expert Validation.} Our human validation team comprises 4 PhD-holding life sciences experts. These experts conduct rigorous review by simultaneously examining both the source scientific papers and the generated questions, ensuring strict verification of knowledge accuracy and appropriateness as evaluation items. They assess whether questions faithfully reflect the paper's content while testing meaningful understanding. This dual-verification approach—checking both scientific correctness and QA suitability—ensures our dataset maintains the highest standards while capturing genuinely challenging, interconnected knowledge that reflects current advances in the field.

Through this rigorous process, we construct $Q_{\text{corpus}} = \{q_1, q_2, ..., q_n\}$ with over 10k validated questions that form the foundation for meaningful graph construction.

\subsection{Graph Construction}
\label{sec:graph-construction}

The core innovation of our approach lies in recovering the latent knowledge structure from the flat question collection. We accomplish this through rigorous graph-based methods that transform $Q_{\text{corpus}}$ into a meaningful evaluation graph. We construct a weighted undirected graph $G = (V, E, w)$ where the structure emerges naturally from the semantic relationships within our question corpus. The node set $V$ corresponds directly to our questions, with $V = Q_{\text{corpus}}$ and $|V| = n$, establishing a one-to-one mapping between graph nodes and evaluation questions.

We first employ an LLM to extract key concepts and entities from each question, generating keyword sets that capture the essential knowledge elements. This extraction process identifies domain-specific terms, scientific concepts, and their semantic roles within the question context. Then we define a hybrid similarity function $\text{sim}: V \times V \rightarrow [0,1]$ that captures both semantic and knowledge-based relationships between questions through a multi-component approach. For questions $v_i$ and $v_j$ with respective keyword sets $K_i$ and $K_j$ extracted by the LLM, we embed all keywords and compute the similarity matrix $S_{ij} \in \mathbb{R}^{|K_i| \times |K_j|}$ where $S_{ij}[p,q] = \cos(\text{emb}(k_p^i), \text{emb}(k_q^j))$. We then compute two complementary similarity components:

\textbf{Bidirectional Matching:} Measures mutual coverage between keyword sets. We first define directional similarities:
\begin{align}
    \text{sim}_{i \rightarrow j} & = \frac{1}{|K_i|}\sum_{p=1}^{|K_i|}\max_q S_{ij}[p,q] \\
    \text{sim}_{j \rightarrow i} & = \frac{1}{|K_j|}\sum_{q=1}^{|K_j|}\max_p S_{ij}[p,q]
\end{align}
where $\text{sim}_{i \rightarrow j}$ measures how well keywords in $K_i$ are covered by $K_j$, and vice versa. The bidirectional similarity is:
\begin{equation}
    \text{sim}_{\text{bi}}(v_i, v_j) = \frac{1}{2}\left(\text{sim}_{i \rightarrow j} + \text{sim}_{j \rightarrow i}\right)
\end{equation}

\textbf{Core Similarity:} Focuses on the strongest connections by selecting top-$k$ matches:
\begin{equation}
    \text{sim}_{\text{core}}(v_i, v_j) = \frac{1}{k}\sum_{l=1}^{k} S^{(l)}_{ij}
\end{equation}
where $S^{(l)}_{ij}$ represents the $l$-th largest element in the flattened similarity matrix, and $k = \min(\kappa, |K_i|, |K_j|)$ with $\kappa$ being a predefined parameter.

The final similarity score combines these two components:
\begin{equation}
    \text{sim}(v_i, v_j) = \gamma \cdot \text{sim}_{\text{core}}(v_i, v_j) + (1-\gamma) \cdot \text{sim}_{\text{bi}}(v_i, v_j)
\end{equation}
where $\gamma \in [0,1]$ controls the balance between core similarity and bidirectional matching.

An edge $(v_i, v_j)$ exists in our graph if and only if $\text{sim}(v_i, v_j) > \theta$, where $\theta$ is an adaptive threshold determined. Each edge carries a weight $w(v_i, v_j) = \text{sim}(v_i, v_j)$, encoding the strength of the relationship between questions. This construction ensures that our graph captures meaningful semantic relationships while avoiding noise from spurious connections.

\section{Graph-Based Model Evaluation}
\label{sec:graph-evaluation}

The graph structure of our benchmark enables fundamentally new approaches to model evaluation. We leverage the evaluation graph topology to assess two core aspects of model performance, its global coherence in understanding life sciences as an integrated knowledge domain and the distribution of its errors across the graph—and develop complementary methods to address these challenges.

\subsection{Global Coherence Score (GCS)}
\fussy
Consider two models achieving identical 75\% accuracy on a life sciences benchmark. Model A correctly answers questions about protein synthesis, translation, and ribosome function as a coherent cluster, while Model B's correct answers scatter randomly—answering about ATP synthesis while missing basic cellular respiration, correctly identifying drug mechanisms while failing on the underlying biochemistry. Traditional accuracy metrics see these models as equivalent, yet any domain expert would immediately recognize Model A's superior understanding. The Global Coherence Score (GCS) captures this phenomenon by recognizing that genuine comprehension creates clusters of consistent performance in the knowledge graph.

For model $M$ evaluated on graph $G = (V, E, w)$, let $\text{Res}_M(v) \in \{0, 1\}$ denote whether $M$ correctly answers question $v$. We compute the neighborhood coherence for each node:
\begin{equation}
    \text{Coherence}_M(v) = \frac{\sum_{u \in N(v)} w(v,u) \cdot \text{Res}_M(u)}{\sum_{u \in N(v)} w(v,u)}
\end{equation}
where $N(v)$ denotes the set of nodes whose distance from $v$ is at most 1. This represents the weighted accuracy within $v$'s semantic neighborhood.

The GCS transforms each binary correct/incorrect outcome into a continuous value that reflects neighborhood support:
\begin{equation}
    \text{GCS}(M) = \frac{1}{|V|} \sum_{v \in V} \text{Res}_M(v) \cdot \text{Coherence}_M(v)
\end{equation}
Although the GCS is strongly correlated with overall accuracy, transforming each binary 0/1 outcome into a weighted value between 0 and 1 allows it to reveal knowledge gaps even among correctly answered nodes with low neighborhood support. This highlights cases where the model may have guessed correctly in isolation, without demonstrating genuine understanding of the surrounding semantic context—making GCS a more diagnostic measure of structured knowledge.

\sloppy
\subsection{Knowledge Balance Score (KBS)}

While GCS measures overall coherence, understanding how consistently a model maintains coherence across different regions of the knowledge graph reveals deeper insights into its knowledge organization patterns. Traditional evaluation metrics fail to capture whether a model exhibits balanced understanding or demonstrates highly variable performance across semantic neighborhoods. We introduce the Knowledge Balance Score (KBS), a novel metric that quantifies the variance in local coherence patterns to assess knowledge stability and balance.

For a given model $M$ evaluated on graph $G = (V, E, w)$, we leverage the neighborhood coherence values $\text{Coherence}_M(v)$ computed in the GCS framework. The KBS is computed as the variance of amplified coherence values across all nodes:
\begin{align}
    \text{KBS}(M) & = \text{Var}(\text{Coherence}_M(v) \times \alpha)                            \\
                  & = \frac{1}{|V|} \sum_{v \in V} (\text{Coherence}_M(v) \times \alpha - \mu)^2
\end{align}
where $\alpha > 1$ is an amplification factor that enhances the distinction between high and low coherence regions, and $\mu = \frac{1}{|V|} \sum_{v \in V} (\text{Coherence}_M(v) \times \alpha)$ is the mean amplified coherence across all nodes.

This variance-based perspective on coherence analysis enables precise identification of whether models suffer from knowledge imbalance or systematic understanding deficits, guiding more effective training and improvement strategies.

\section{Experiments}
\label{sec:experiments}

\subsection{Experimental Setup}
\label{sec:experimental-setup}

\textbf{Benchmark Details.} Our dataset construction pipeline generated high-quality questions from recent peer-reviewed literature spanning three major life sciences domains. For each subdomain, the corresponding graph is a subgraph of the global evaluation graph. Table~\ref{tab:benchmark-details} presents the structural characteristics of our constructed evaluation graph. More details can be found in Appendix~\ref{ap1}.

\begin{table}[h]
    \centering
    \caption{LG-Bench: Graph-Structured Benchmark Statistics}
    \begin{tabular}{cccc}
        \toprule
        \textbf{Domain}  & \textbf{Questions (Nodes)} & \textbf{Semantic Edges} & \textbf{Avg. Degree} \\
        \midrule
        Biology          & 3,630                      & 521,890                 & 287.50               \\
        Chemistry        & 3,360                      & 318,763                 & 189.74               \\
        Medicine         & 3,210                      & 437,442                 & 272.55               \\
        \midrule
        \textbf{Overall} & \textbf{10,200}            & \textbf{2,990,277}      & \textbf{586.32}      \\
        \bottomrule
    \end{tabular}%
    \label{tab:benchmark-details}
\end{table}

\textbf{Models.} We evaluate a comprehensive set of large language models spanning different scales, architectures, and specializations to provide a thorough assessment of capabilities. In the case of open-source models , we include the Qwen2.5 series (0.5B/7B/14B/32B/72B) \citep{qwen2.5}, the Llama family \citep{touvron2023llama,meta2024llama3,dubey2024llama} including Llama-7B, Llama-3-8b, Llama-3.3-70B, Llama-3.1-405B, and Llama-4-scout. We also evaluate the GLM-4 series (9B/32B)  \citep{glm2024chatglm}, Gemma-3-1B  \citep{gemma_2025}, and DeepSeek-v3  \citep{deepseekai2024deepseekv3technicalreport}. For domain-specific models, we include BioMedLM-2.7B  \citep{bolton2024biomedlm}, Medicine-LLM-7B  \citep{cheng2024adapting}, HuatuoGPT family  \citep{zhang2023huatuogpt,chen2023huatuogpt}, and Llama3-Med42-8B  \citep{med42v2}, which are specifically fine-tuned for life sciences applications.

For closed-source commercial models, we utilize the GPT series including GPT-3.5, GPT-4o-mini, GPT-4o, OpenAI o3, OpenAI o4-mini an GPT-5  \citep{openai2023a,openai2023b, singh2025openaigpt5card}, the Claude family with Claude-3.5-sonnet and Claude-3.7-sonnet  \citep{anthropic_claude35,anthropic_claude37}, and the Grok series (Grok-2/Grok-3)  \citep{xai2025grok3}. All models are evaluated using identical prompting strategies and evaluation protocols to ensure fair comparison.

\subsection{Results}
\label{sec:results}

Table~\ref{tab:main-results} presents comprehensive evaluation results across all models on LG-Bench using our graph-based metrics. All accuracy (Acc) and Global Coherence Score (GCS) values are reported as percentages. For KBS computation, we set the amplification parameter $\alpha = 100$ to enhance the distinction between high and low coherence regions. All experiments were conducted three times and averaged, with more experimental settings available in Appendix~B. Our evaluation demonstrates significant performance variations across models, with overall accuracy ranging from 23.93\% (BioMedLM-2.7B) to 91.00\% (GPT-5), GCS spanning 5.67\% to 83.10\%, and KBS values varying from 4.40 to 20.00. Notably, our graph-based metrics reveal substantial differences between models that are not captured by traditional accuracy alone, effectively distinguishing models with coherent knowledge organization from those exhibiting scattered understanding patterns across the life sciences domain.

\begin{wrapfigure}[29]{R}{.5\linewidth}
    \vspace{-\baselineskip}
    \centering
    \includegraphics[width=0.95\linewidth]{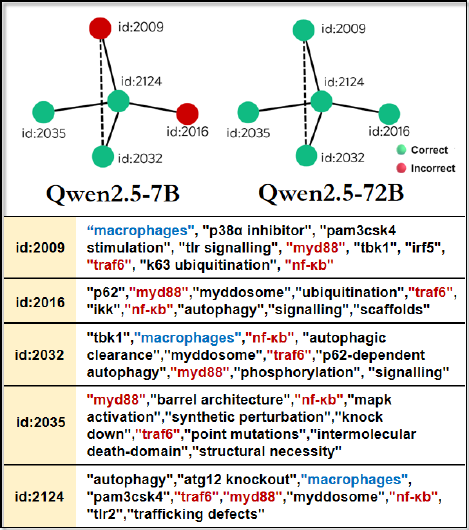}
    \caption{Qwen2.5 results within a local graph structure. Qwen2.5-7B failed on many questions in the neighborhood-level evaluation. Red keywords indicate areas of severe knowledge weakness, while blue keywords denote relatively weak points. In contrast, Qwen2.5-72B correctly answered all relevant questions in the neighborhood test, demonstrating a relatively good grasp of the associated knowledge.}
    \label{fig:case2}
\end{wrapfigure}

As shown in Fig.\ref{fig:case2}, we compare the responses of Qwen2.5 within a local graph structure and observe a clear lack of understanding of certain concepts in Qwen2.5-7B, whereas Qwen2.5-72B demonstrates a clear and solid grasp of these knowledge points. A more detailed analysis is provided in Appendix~\ref{ap2}.

\long\def\MainResultsTable{%
\begin{table}[!t]
    \centering
    \caption{Evaluation Results on LG-Bench: Acc denotes traditional accuracy (in \%), GCS measures global coherence (in \%), with higher scores indicating stronger capabilities. We use \textbf{bold} to highlight the best-performing model in each domain, and \textit{italics} to indicate the second-best. KBS quantifies knowledge balance, and higher scores indicate a more uneven distribution. For hyperparameter settings, we set $\gamma = 0.6$ and $\theta = 0.4$.}
    \label{tab:main-results}
    \setlength{\tabcolsep}{3pt}
    \renewcommand{\arraystretch}{0.82}
    \resizebox{\linewidth}{!}{%
    \begin{tabular}{ccccccccccccc}
        \toprule
        \multicolumn{1}{c|}{\multirow{2}{*}{Model}}   & \multicolumn{3}{c|}{Medicine} & \multicolumn{3}{c|}{Biology} & \multicolumn{3}{c|}{Chemistry} & \multicolumn{3}{c}{Overall}                                                                                                                                                        \\ \cmidrule{2-13}
        \multicolumn{1}{c|}{}                         & Acc                           & GCS                          & \multicolumn{1}{c|}{KBS}       & Acc                         & GCS            & \multicolumn{1}{c|}{KBS}   & Acc            & GCS            & \multicolumn{1}{c|}{KBS}   & Acc            & GCS            & KBS   \\ \midrule
        \multicolumn{13}{c}{Open-Weight LLM (Scale\textless{}10B)}                                                                                                                                                                                                                                                                         \\ \midrule
        \multicolumn{1}{c|}{Qwen2.5-0.5B        }     & 37.07                         & 13.96                        & \multicolumn{1}{c|}{22.19}     & 38.10                       & 14.83          & \multicolumn{1}{c|}{26.34} & 37.44          & 14.35          & \multicolumn{1}{c|}{36.52} & 37.56          & 14.39          & 16.11 \\
        \multicolumn{1}{c|}{Gemma-3-1b          }     & 51.56                         & 26.73                        & \multicolumn{1}{c|}{29.42}     & 55.45                       & 30.71          & \multicolumn{1}{c|}{21.97} & 52.65          & 28.20          & \multicolumn{1}{c|}{39.34} & 53.30          & 28.62          & 17.94 \\
        \multicolumn{1}{c|}{BioMedLM-2.7B       }     & 24.14                         & 5.80                         & \multicolumn{1}{c|}{12.85}     & 23.80                       & 5.65           & \multicolumn{1}{c|}{9.03}  & 23.87          & 5.56           & \multicolumn{1}{c|}{13.69} & 23.93          & 5.67           & 5.28  \\
        \multicolumn{1}{c|}{Llama-7B            }     & 30.06                         & 9.19                         & \multicolumn{1}{c|}{13.62}     & 30.91                       & 9.60           & \multicolumn{1}{c|}{14.41} & 32.20          & 10.47          & \multicolumn{1}{c|}{19.14} & 31.07          & 9.76           & 6.80  \\
        \multicolumn{1}{c|}{Medicine-LLM-7B     }     & 38.72                         & 15.23                        & \multicolumn{1}{c|}{22.23}     & 34.88                       & 12.26          & \multicolumn{1}{c|}{17.23} & 35.57          & 12.58          & \multicolumn{1}{c|}{23.87} & 36.31          & 13.29          & 10.65 \\
        \multicolumn{1}{c|}{HuatuoGPT-7B        }     & 28.75                         & 8.36                         & \multicolumn{1}{c|}{15.94}     & 29.20                       & 8.50           & \multicolumn{1}{c|}{10.31} & 28.42          & 8.13           & \multicolumn{1}{c|}{12.97} & 28.80          & 8.33           & 5.96  \\
        \multicolumn{1}{c|}{HuatuoGPT2-7B       }     & 61.06                         & 38.09                        & \multicolumn{1}{c|}{33.85}     & 62.42                       & 39.36          & \multicolumn{1}{c|}{20.89} & 58.36          & 34.41          & \multicolumn{1}{c|}{44.45} & 60.66          & 37.39          & 20.00 \\
        \multicolumn{1}{c|}{Qwen2.5-7B          }     & 78.38                         & 62.15                        & \multicolumn{1}{c|}{29.17}     & 80.03                       & 64.24          & \multicolumn{1}{c|}{21.58} & 76.76          & 59.21          & \multicolumn{1}{c|}{25.74} & 78.43          & 62.00          & 15.19 \\
        \multicolumn{1}{c|}{Llama-3-8b          }     & 72.77                         & 54.11                        & \multicolumn{1}{c|}{35.65}     & 75.67                       & 57.64          & \multicolumn{1}{c|}{17.94} & 72.53          & 53.16          & \multicolumn{1}{c|}{27.97} & 73.73          & 55.04          & 14.62 \\
        \multicolumn{1}{c|}{Llama3-Med42-8B     }     & 78.10                         & 61.60                        & \multicolumn{1}{c|}{25.61}     & 77.38                       & 60.01          & \multicolumn{1}{c|}{18.22} & 75.00          & 56.71          & \multicolumn{1}{c|}{30.78} & 76.82          & 59.43          & 13.56 \\
        \multicolumn{1}{c|}{GLM-4-9b            }     & 76.67                         & 59.61                        & \multicolumn{1}{c|}{30.59}     & 77.47                       & 59.82          & \multicolumn{1}{c|}{25.99} & 75.45          & 57.50          & \multicolumn{1}{c|}{36.28} & 76.55          & 58.97          & 18.05 \\
        \midrule
        \multicolumn{13}{c}{Open-Weight LLM (Scale\textgreater{}10B)}                                                                                                                                                                                                                                                                      \\ \midrule
        \multicolumn{1}{c|}{Qwen2.5-14B         }     & 83.02                         & 69.62                        & \multicolumn{1}{c|}{20.67}     & 85.26                       & 72.72          & \multicolumn{1}{c|}{18.54} & 82.62          & 68.64          & \multicolumn{1}{c|}{21.91} & 83.69          & 70.42          & 11.73 \\
        \multicolumn{1}{c|}{Qwen2.5-32B         }     & 84.70                         & 72.55                        & \multicolumn{1}{c|}{21.01}     & 86.06                       & 74.34          & \multicolumn{1}{c|}{15.02} & 83.75          & 70.38          & \multicolumn{1}{c|}{20.78} & 84.87          & 72.56          & 11.07 \\
        \multicolumn{1}{c|}{GLM-4-32B           }     & 78.94                         & 63.32                        & \multicolumn{1}{c|}{29.30}     & 80.44                       & 64.82          & \multicolumn{1}{c|}{22.22} & 78.30          & 61.77          & \multicolumn{1}{c|}{29.31} & 79.26          & 63.33          & 15.28 \\
        \multicolumn{1}{c|}{Llama-3.1-70B           } & 72.77                         & 69.51                        & \multicolumn{1}{c|}{24.30}     & 83.14                       & 69.19          & \multicolumn{1}{c|}{19.55} & 71.67          & 66.78          & \multicolumn{1}{c|}{24.67} & 82.54          & 68.58          & 12.70 \\
        \multicolumn{1}{c|}{Qwen2.5-72B         }     & 86.26                         & 75.12                        & \multicolumn{1}{c|}{16.53}     & 86.97                       & 75.68          & \multicolumn{1}{c|}{15.45} & 84.85          & 72.44          & \multicolumn{1}{c|}{23.04} & 86.05          & 74.52          & 10.26 \\
        \multicolumn{1}{c|}{Llama-3.1-405b      }     & 88.07                         & 78.13                        & \multicolumn{1}{c|}{13.98}     & 88.26                       & 78.01          & \multicolumn{1}{c|}{10.88} & 87.14          & 76.18          & \multicolumn{1}{c|}{14.98} & 87.83          & 77.51          & 7.19  \\
        \multicolumn{1}{c|}{Llama-4-scout       }     & 78.69                         & 62.61                        & \multicolumn{1}{c|}{32.32}     & 81.79                       & 66.95          & \multicolumn{1}{c|}{25.66} & 80.27          & 64.79          & \multicolumn{1}{c|}{28.76} & 80.31          & 64.89          & 18.65 \\
        \multicolumn{1}{c|}{Deepseek-v3         }     & 87.17                         & 76.57                        & \multicolumn{1}{c|}{14.76}     & 87.96                       & 77.51          & \multicolumn{1}{c|}{11.63} & 86.90          & 75.70          & \multicolumn{1}{c|}{15.09} & 87.36          & 76.70          & 7.50  \\
        \midrule
        \multicolumn{13}{c}{Closed-Weight LLM}                                                                                                                                                                                                                                                                                             \\ \midrule
        \multicolumn{1}{c|}{Grok-2              }     & 88.54                         & 78.71                        & \multicolumn{1}{c|}{16.61}     & 89.01                       & 79.30          & \multicolumn{1}{c|}{11.15} & 87.23          & 76.36          & \multicolumn{1}{c|}{17.60} & 88.27          & 78.20          & 7.78  \\
        \multicolumn{1}{c|}{Grok-3              }     & 86.29                         & 75.06                        & \multicolumn{1}{c|}{14.01}     & 84.38                       & 71.75          & \multicolumn{1}{c|}{29.89} & 85.77          & 73.75          & \multicolumn{1}{c|}{16.68} & 85.44          & 73.54          & 10.40 \\
        \multicolumn{1}{c|}{Claude-3.5-sonnet   }     & 88.91                         & 79.43                        & \multicolumn{1}{c|}{12.51}     & 88.65                       & 78.61          & \multicolumn{1}{c|}{10.25} & 86.93          & 76.03          & \multicolumn{1}{c|}{18.69} & 88.17          & 78.09          & 7.56  \\
        \multicolumn{1}{c|}{Claude-3.7-sonnet   }     & 88.41                         & 78.76                        & \multicolumn{1}{c|}{15.91}     & 89.04                       & 79.29          & \multicolumn{1}{c|}{9.00}  & 87.83          & 77.27          & \multicolumn{1}{c|}{14.71} & 88.44          & 78.52          & 6.97  \\
        \multicolumn{1}{c|}{GPT-3.5             }     & 76.42                         & 59.35                        & \multicolumn{1}{c|}{27.79}     & 77.02                       & 59.67          & \multicolumn{1}{c|}{23.01} & 74.55          & 56.38          & \multicolumn{1}{c|}{35.35} & 76.02          & 58.47          & 16.31 \\
        \multicolumn{1}{c|}{GPT-4o-mini         }     & 82.21                         & 68.77                        & \multicolumn{1}{c|}{25.94}     & 84.41                       & 71.13          & \multicolumn{1}{c|}{14.91} & 81.70          & 67.26          & \multicolumn{1}{c|}{26.46} & 82.82          & 69.19          & 12.68 \\
        \multicolumn{1}{c|}{GPT-4o              }     & 88.47                         & 78.99                        & \multicolumn{1}{c|}{15.08}     & \textit{89.70}              & \textit{80.44} & \multicolumn{1}{c|}{10.62} & 88.01          & 77.79          & \multicolumn{1}{c|}{13.29} & 88.75          & 79.19          & 6.22  \\

        \multicolumn{1}{c|}{OpenAI o3}                & \textit{90.00}                & \textit{81.57}               & \multicolumn{1}{c|}{10.65}     & 89.26                       & 79.84          & \multicolumn{1}{c|}{9.94}  & \textit{89.05} & \textit{79.56} & \multicolumn{1}{c|}{11.60} & 89.42          & 80.34          & 5.83  \\
        \multicolumn{1}{c|}{OpenAI o4-mini}           & 89.75                         & 81.09                        & \multicolumn{1}{c|}{10.69}     & \textit{89.89}              & \textit{81.14} & \multicolumn{1}{c|}{7.22}  & 88.87          & 79.32          & \multicolumn{1}{c|}{9.46}  & \textit{89.51} & \textit{80.59} & 4.45  \\
        \multicolumn{1}{c|}{GPT-5}                    & \textbf{91.31}                & \textbf{83.79}               & \multicolumn{1}{c|}{9.10}      & \textbf{91.05}              & \textbf{83.10} & \multicolumn{1}{c|}{7.88}  & \textbf{90.65} & \textbf{82.10} & \multicolumn{1}{c|}{8.66}  & \textbf{91.00} & \textbf{83.10} & 4.40  \\

        \bottomrule
    \end{tabular}
    }%
\end{table}%
}

\textbf{Open-Weight LLMs \textit{vs.} Closed-Weight LLMs.} The result highlights a striking contrast between open and closed models when confronted with LG-Bench, whose items deliberately target clinical guidelines, drugs, and biomolecular findings released recently. Early open models such as HuatuoGPT, BioMedLM, and Llama-7B reach only 24–31\% accuracy, revealing substantial blind spots for the newest knowledge. In contrast, iterative open-source versions like HuatuoGPT2-7B, Llama-3-8B, and Llama3-Med42-8B boost the same metric to 60–78\% while staying under 10B parameters. These improvements attest to the compounding benefits of an open ecosystem, where community-contributed data refreshes and lightweight domain fine-tuning shorten the model–data–task loop.
Closed-source systems still occupy the top tier: GPT-5, OpenAI o3 and OpenAI o4-mini surpass 89\% overall accuracy, reflecting the value of proprietary corpora and extensive RLHF \citep{christiano2017deep}. Nevertheless, the gap is narrowing; fully open models like Qwen2.5-72B and DeepSeek-v3 trail GPT series by less than 4 pp. In sum, while closed models presently set the performance ceiling, rapid, community-driven iteration is continuously raising the open-source floor, accelerating progress for medical reasoning at large.

\textbf{Model Scale Effects.} Our empirical analysis confirms a clear parameter--performance scaling law. Sub-10B models deliver only moderate results, with accuracies below 80\% and GCS under 62\%; for instance, Qwen2.5-0.5B attains merely 37.56\% accuracy and a GCS of 14.39\%. Once the parameter count exceeds 10B, every metric improves sharply. Within the Qwen2.5 family, accuracy and GCS rise from 83.69\% and 70.42\% at 14B to 84.87\% and 72.56\% at 32B, and further to 86.05\% and 74.52\% at 72B. Knowledge balance benefits from scaling as well: the KBS drops monotonically from 16.11 at 0.5B to 11.73 at 14B, 11.07 at 32B, and 10.26 at 72B, indicating progressively more coherent knowledge organization. The Llama series exhibits the same scaling advantages, underscoring the generality of these trends across model families. The Coherence Score distrubution of Qwen2.5 family is shown as Fig. ~\ref{fig:30}.

\long\def\CoherenceHeatmapFigures{%
\begin{figure}[!ht]
    \centering
    \begin{minipage}[t]{0.49\textwidth}
        \vspace{0pt}
        \centering
        \begin{subfigure}[t]{0.49\linewidth}
            \includegraphics[width=\linewidth]{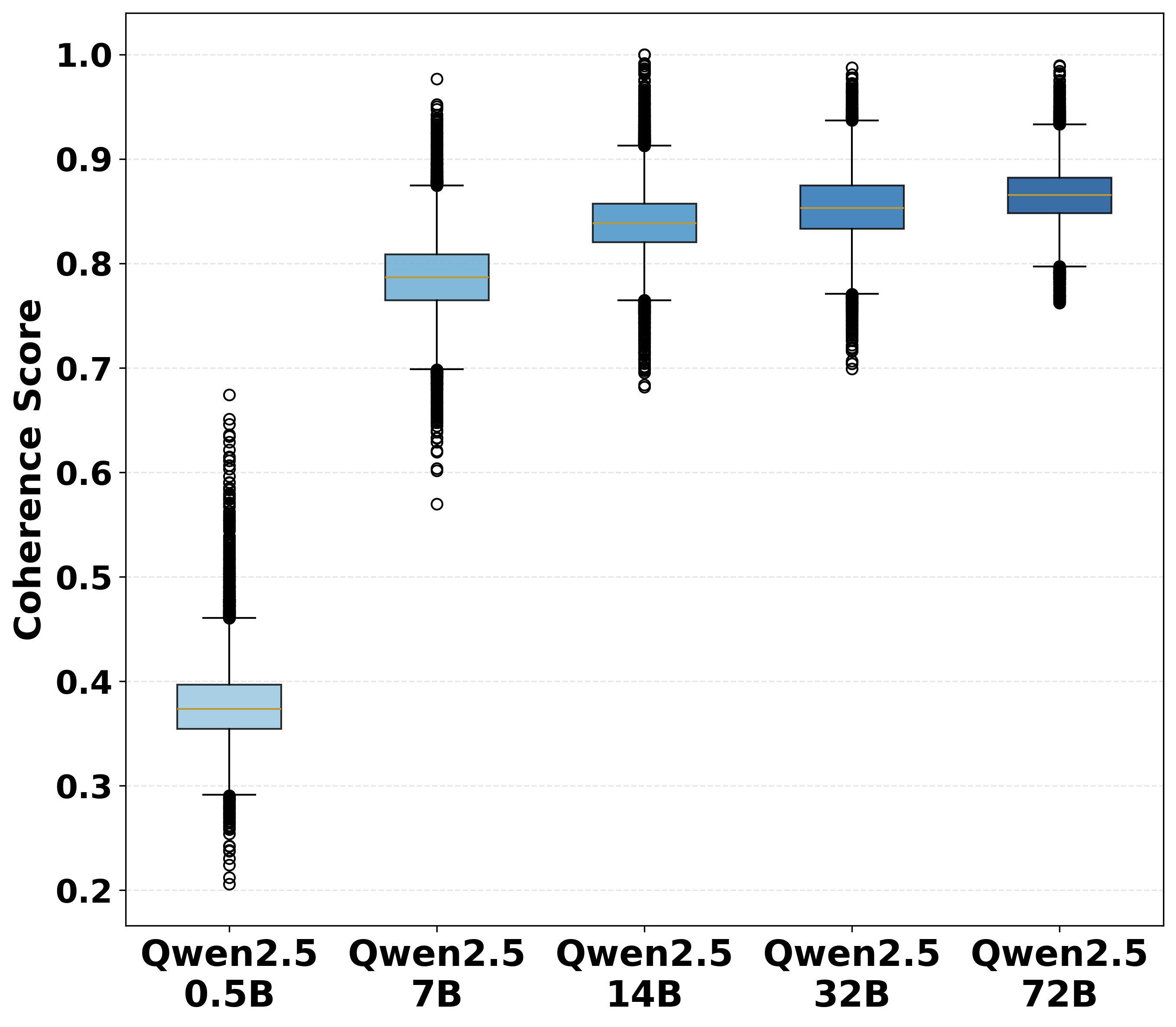}
            \caption{Qwen2.5}
            \label{fig:img30}
        \end{subfigure}
        \hfill
        \begin{subfigure}[t]{0.49\linewidth}
            \includegraphics[width=\linewidth]{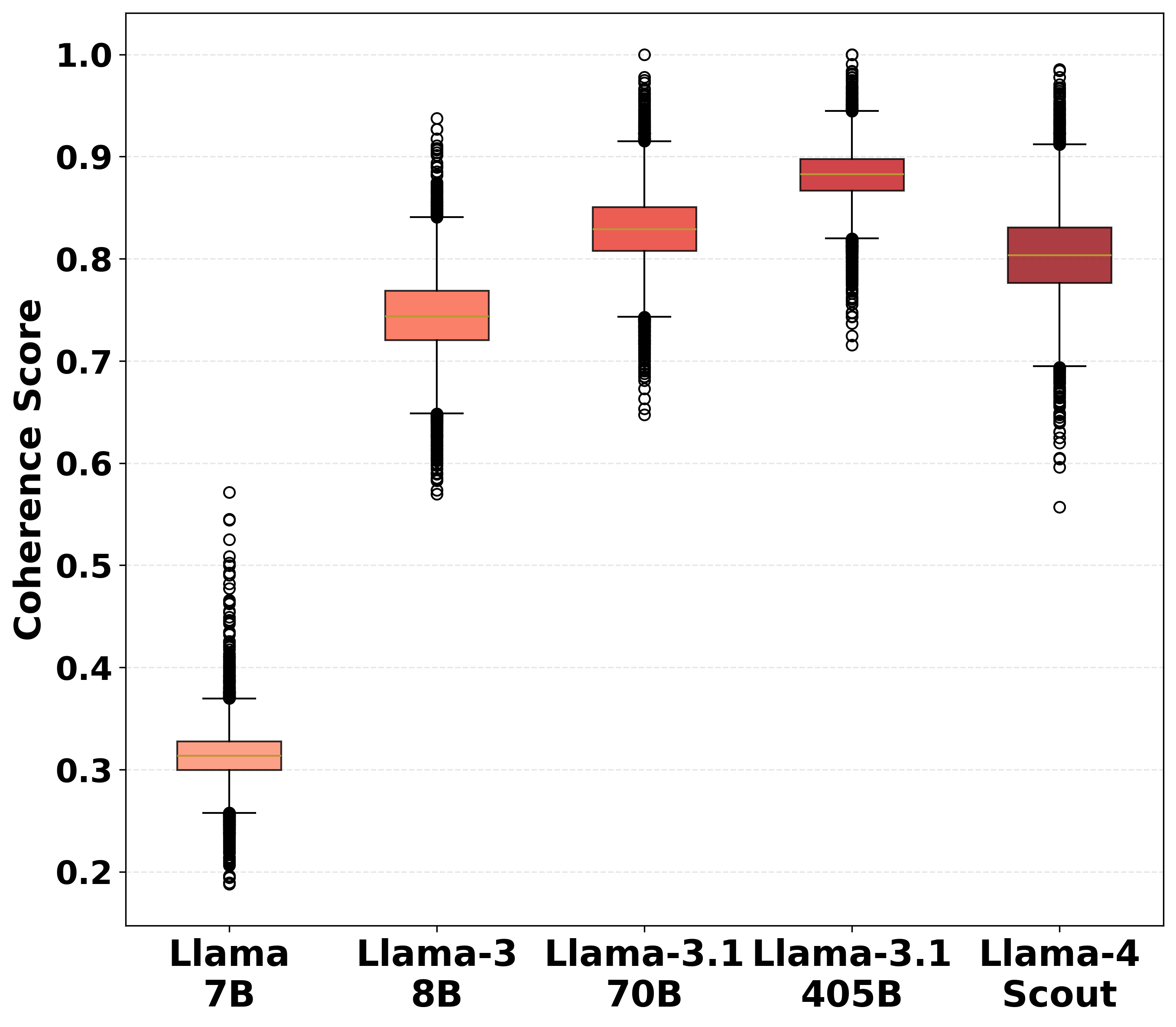}
            \caption{Llama}
            \label{fig:img40}
        \end{subfigure}
        \caption{Coherence Score distribution of Qwen2.5 and Llama family. As model size increases, its overall capability in the life sciences domain gradually improves, while its knowledge becomes more evenly distributed.}
        \label{fig:30}
    \end{minipage}
    \hfill
    \begin{minipage}[t]{0.49\textwidth}
        \vspace{0pt}
        \centering
        \begin{subfigure}[t]{0.49\linewidth}
            \includegraphics[width=\linewidth]{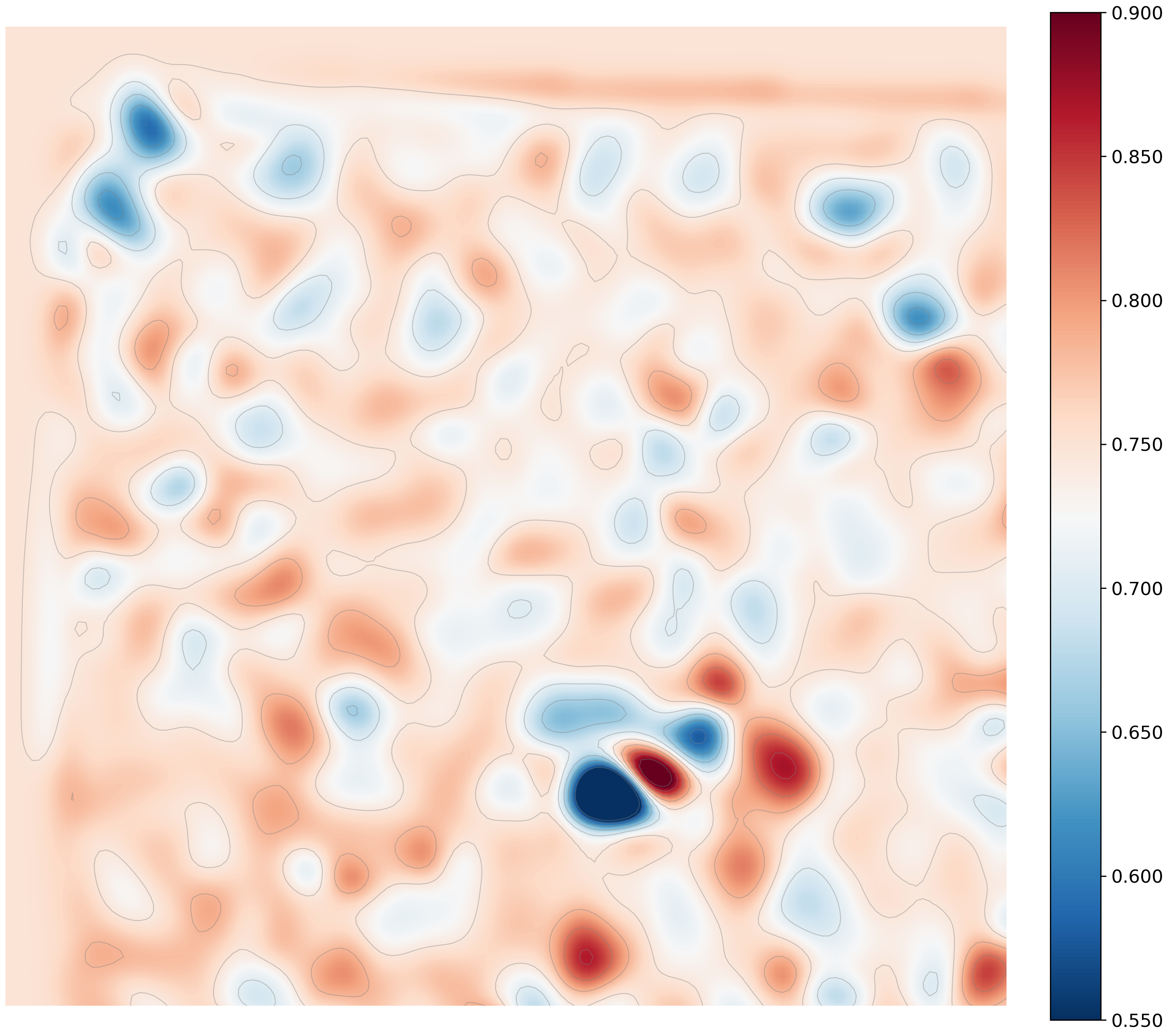}
            \caption{Llama-3-8b}
            \label{fig:img3}
        \end{subfigure}
        \hfill
        \begin{subfigure}[t]{0.49\linewidth}
            \includegraphics[width=\linewidth]{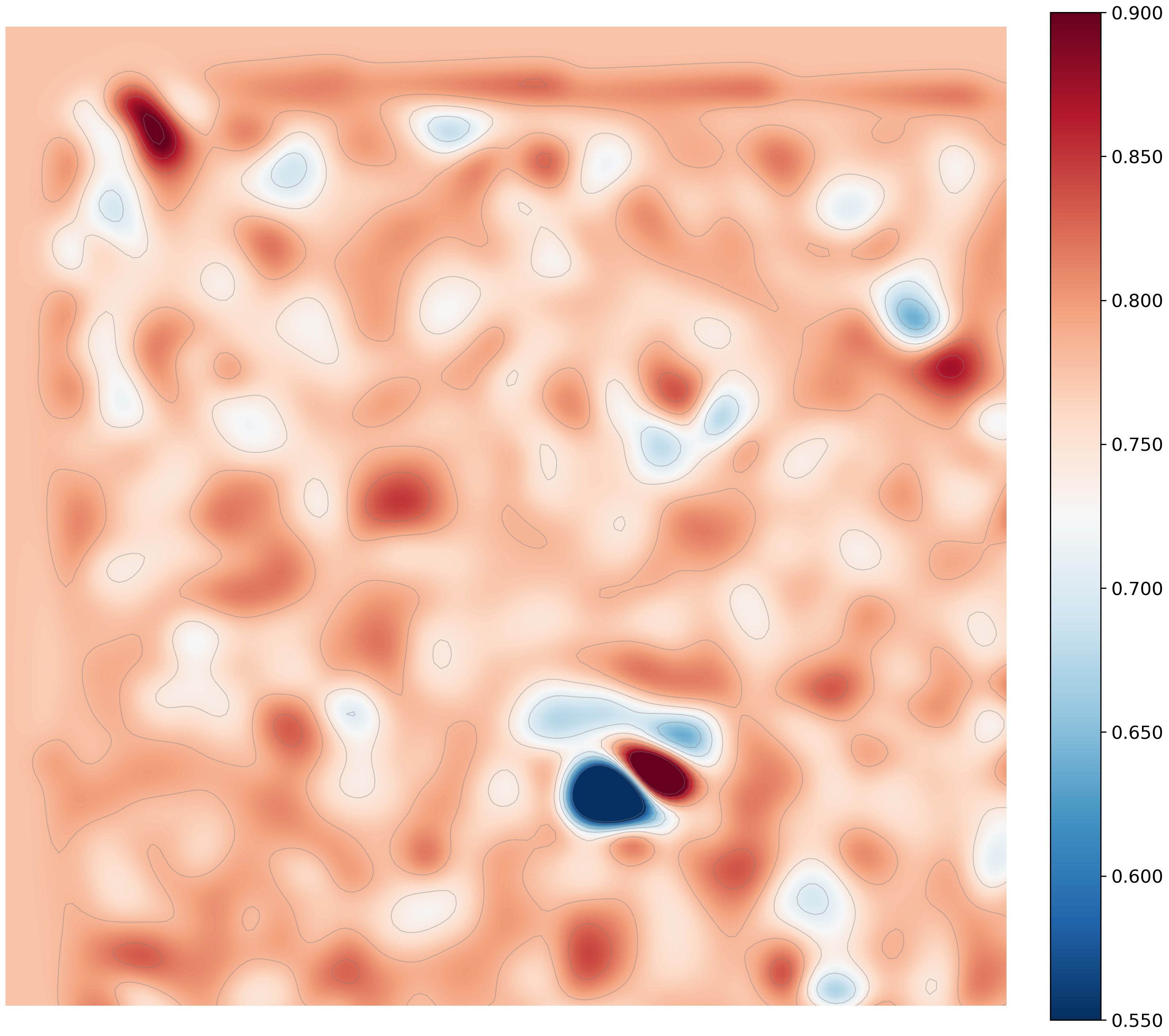}
            \caption{Llama3-Med42-8B}
            \label{fig:img4}
        \end{subfigure}
        \caption{Heatmap visualization of the LG-Bench. Each cell represents a node's Coherence Score. Redder regions indicate higher coherence, while bluer areas indicate lower coherence.}
        \label{fig:3}
    \end{minipage}
\end{figure}%
}

\textbf{Impact of Domain-Specific Training.} Our graph-based metrics provide a detailed lens through which to analyze the effects of domain-specific fine-tuning. A direct comparison between the generalist Llama-3-8b and its domain-adapted counterpart, Llama3-Med42-8B, reveals that specialization does more than just increase accuracy---it fundamentally reshapes the model's knowledge structure. We embed each knowledge graph with \emph{node2vec}  \citep{grover2016node2vec} into a 2-D manifold and plot Coherence Score heatmaps, shown as Fig.~\ref{fig:3}. Although the overall surface becomes brighter—corroborating the aggregate metric improvements—an inverted pattern emerges in certain regions, revealing a migration of knowledge density rather than a uniform amplification.

In the target domain of Medicine, Llama3-Med42-8B shows a substantial improvement in its  GCS, rising to 61.60\% from the base model's 54.11\%.  More revealing is the impact on the  KBS. The KBS for Llama3-Med42-8B in Medicine drops sharply to 25.61 from the base model's 35.65. This significant reduction in variance demonstrates that the fine-tuning not only enhanced knowledge but also homogenized it, leading to a more evenly distributed and balanced understanding across different medical topics. However, this specialization comes with trade-offs that metrics can precisely identify. While its GCS shows minor improvements in the related field of Biology, Llama3-Med42-8B's KBS in Chemistry increases to 30.78, higher than the base model's 27.97. This suggests that the intense focus on medicine may have inadvertently created a more imbalanced and "spiky" knowledge representation in the less related chemical domain. This demonstrates that domain adaptation can reshape knowledge organization both positively and negatively across different semantic regions, a critical insight that flat accuracy metrics would completely miss.

\MainResultsTable
\FloatBarrier

\subsection{Robustness and Human Alignment}
\label{sec:robustness_main}

The reliability of our graph-based metrics, GCS and KBS, depends on the underlying graph topology, which is primarily governed by the similarity threshold $\theta$ and the connectivity parameter $\gamma$. While LG-Bench is established with default configurations of $\theta = 0.4$ and $\gamma = 0.6$, we conducted extensive sensitivity analyses to ensure that our evaluation conclusions are not artifacts of specific hyperparameter choices.

We first investigated the impact of the similarity threshold $\theta$, which serves as the primary determinant of graph density. By increasing $\theta$ to 0.45, as shown in Table~\ref{tab:robustness_gamma_ablation}, and further to 0.5 and 0.6 in Appendix~\ref{Robustness}, we generated increasingly sparse graph structures. Our results demonstrate that GCS exhibits remarkable numerical stability even as the graph topology becomes more restrictive. More importantly, while the absolute magnitude of KBS naturally scales with changes in sparsity due to its variance-based calculation, the relative rank ordering of models remains strictly consistent across all tested $\theta$ values.

\CoherenceHeatmapFigures
\FloatBarrier

\begin{wraptable}[27]{r}{.60\linewidth}
    \vspace{-\baselineskip}
    \centering
    \caption{Ablation Study on Graph Connectivity Parameters. We study metric sensitivity to graph structure by varying $\gamma$ at a fixed $\theta = 0.45$ (originally $\theta = 0.4$, $\gamma = 0.6$). Results show \textbf{GCS} is stable, indicating global coherence is not an artifact of graph density. \textbf{KBS} scales with sparsity, but model rank ordering remains consistent across domains. Table (a) reports $\gamma = 0.4$; Table (b) $\gamma = 0.6$.}
    \label{tab:robustness_gamma_ablation}
    \scriptsize
    \setlength{\tabcolsep}{1.5pt}
    \renewcommand{\arraystretch}{0.88}
    \textbf{(a) Setting $\gamma = 0.4$ (Avg. Degree $\approx$ 249.61)} \\
    \resizebox{\linewidth}{!}{
        \begin{tabular}{l|cc|cc|cc|cc}
            \toprule
            \multirow{2}{*}{\textbf{Model}} & \multicolumn{2}{c|}{\textbf{Medicine}} & \multicolumn{2}{c|}{\textbf{Biology}} & \multicolumn{2}{c|}{\textbf{Chemistry}} & \multicolumn{2}{c}{\textbf{Overall}}                                                             \\
            \cmidrule(lr){2-3} \cmidrule(lr){4-5} \cmidrule(lr){6-7} \cmidrule(lr){8-9}
                                            & \textbf{GCS}                           & \textbf{KBS}                          & \textbf{GCS}                            & \textbf{KBS}                         & \textbf{GCS} & \textbf{KBS} & \textbf{GCS} & \textbf{KBS} \\
            \midrule
            Qwen2.5-7B                      & 62.59                                  & 60.38                                 & 64.29                                   & 40.12                                & 59.29        & 48.07        & 62.20        & 29.83        \\
            Llama-3-8B                      & 54.41                                  & 69.05                                 & 57.68                                   & 32.90                                & 53.21        & 52.91        & 55.25        & 29.14        \\
            Llama3-Med42-8B                 & 61.79                                  & 55.59                                 & 60.17                                   & 35.06                                & 56.82        & 55.66        & 59.59        & 27.72        \\
            Qwen2.5-14B                     & 69.84                                  & 43.47                                 & 72.75                                   & 36.85                                & 68.81        & 38.89        & 70.56        & 22.18        \\
            Qwen2.5-32B                     & 72.79                                  & 44.72                                 & 74.41                                   & 29.19                                & 70.58        & 38.19        & 72.78        & 21.94        \\
            Llama-3.1-70B                   & 69.74                                  & 48.59                                 & 69.25                                   & 38.64                                & 66.95        & 45.48        & 68.75        & 25.85        \\
            Qwen2.5-72B                     & 75.37                                  & 38.76                                 & 75.76                                   & 29.24                                & 72.67        & 41.59        & 74.73        & 20.73        \\
            \bottomrule
        \end{tabular}%
    }
    \vspace{0.5em}
    \textbf{(b) Setting $\gamma = 0.6$ (Avg. Degree $\approx$ 257.42)} \\
    \resizebox{\linewidth}{!}{
        \begin{tabular}{l|cc|cc|cc|cc}
            \toprule
            \multirow{2}{*}{\textbf{Model}} & \multicolumn{2}{c|}{\textbf{Medicine}} & \multicolumn{2}{c|}{\textbf{Biology}} & \multicolumn{2}{c|}{\textbf{Chemistry}} & \multicolumn{2}{c}{\textbf{Overall}}                                                             \\
            \cmidrule(lr){2-3} \cmidrule(lr){4-5} \cmidrule(lr){6-7} \cmidrule(lr){8-9}
                                            & \textbf{GCS}                           & \textbf{KBS}                          & \textbf{GCS}                            & \textbf{KBS}                         & \textbf{GCS} & \textbf{KBS} & \textbf{GCS} & \textbf{KBS} \\
            \midrule
            Qwen2.5-7B                      & 62.05                                  & 61.25                                 & 64.72                                   & 40.89                                & 59.81        & 48.94        & 62.68        & 30.56        \\
            Llama-3-8B                      & 54.86                                  & 70.01                                 & 58.15                                   & 33.72                                & 53.74        & 53.85        & 55.78        & 29.96        \\
            Llama3-Med42-8B                 & 62.31                                  & 55.52                                 & 60.64                                   & 35.91                                & 57.38        & 56.73        & 60.12        & 28.45        \\
            Qwen2.5-14B                     & 70.29                                  & 44.38                                 & 73.21                                   & 37.69                                & 69.35        & 39.82        & 71.09        & 22.91        \\
            Qwen2.5-32B                     & 73.26                                  & 45.67                                 & 74.94                                   & 30.03                                & 71.13        & 39.14        & 73.32        & 22.68        \\
            Llama-3.1-70B                   & 70.18                                  & 49.51                                 & 69.76                                   & 39.58                                & 67.49        & 46.42        & 69.28        & 26.67        \\
            Qwen2.5-72B                     & 75.83                                  & 39.68                                 & 76.28                                   & 30.16                                & 73.21        & 42.54        & 75.27        & 21.55        \\
            \bottomrule
        \end{tabular}%
    }
    \vspace{-0.5\baselineskip}
\end{wraptable}

This rank-order consistency confirms that our framework captures intrinsic differences in knowledge organization rather than reflecting noise in the graph construction process.

Furthermore, we analyzed the influence of the hybrid similarity parameter $\gamma$ by comparing settings of $\gamma = 0.4$ and $\gamma = 0.6$ at a fixed threshold. As illustrated in Table~\ref{tab:robustness_gamma_ablation}, the variations in the weighting between core and bidirectional similarity do not disrupt the performance hierarchy. Models of different scales and specialized domain-adapted models consistently maintain their respective positions in the ranking, proving that the diagnostic capacity of LG-Bench, particularly its ability to identify imbalanced or spiky knowledge distributions, is robust to variations in semantic connectivity.

To confirm that these graph-based scores truly reflect actual reasoning capabilities, we conducted a human verification study on a subset of 450 questions. Human evaluators performed a subjective assessment of model-generated open-ended responses, focusing on the depth and accuracy of the underlying reasoning. We then computed the Spearman rank correlation between human ratings and our automated metrics. The analysis revealed a very high correlation, with coefficients reaching 0.94 for GCS and 0.90 for KBS. These results indicate that the coherence and balance captured by our graph structure are highly aligned with human expert perception of scientific understanding, providing strong empirical evidence for the validity of our diagnostic framework.

\section{Conclusion}

In this work, we introduced LG-Bench, the first graph-structured benchmark designed to address the fundamental limitations of traditional evaluations in the life sciences. By modeling knowledge as a graph and introducing novel coherence-based metrics, we provide the community with the tools to move beyond simple accuracy and assess the depth of a model's scientific reasoning.

\section*{Acknowledgments}
This work was supported by the National Natural Science Foundation of China (Grant Nos. 62501337, 72293585, and 72293580), the National Science and Technology Major Project of New Generation Artificial Intelligence (Grant No. 2025ZD0124104), and the Shanghai Municipal Special Program for Basic Research on General AI Foundation Models (Grant No. 2025SHZDZX025D09). This work was conducted in collaboration with Shanghai Artificial Intelligence Laboratory.

\bibliographystyle{ACM-Reference-Format}
\bibliography{sample-base}


\appendix
\section{Appendix}
\subsection{Limitations and Broader Impacts}
\label{app:limitations_impacts}

\textbf{Limitations.} LG-Bench is built from recent life-sciences literature and currently focuses on medicine, biology, and chemistry, so it may not cover all subfields or all forms of scientific reasoning. Its graph structure also depends on LLM-based concept extraction and semantic similarity, which may introduce noisy or missing edges despite the robustness checks. Therefore, LG-Bench should be viewed as a diagnostic benchmark rather than a direct guarantee of reliable real-world biomedical deployment.

\textbf{Broader Impacts and Future Work.} We believe this evaluation framework can help guide LLMs beyond pattern matching toward more coherent scientific understanding, ultimately supporting discovery and innovation across the life sciences. At the same time, benchmark scores should not be used as the sole basis for high-stakes biomedical decisions, since overinterpreting model performance may create safety risks. In future work, we will continuously update the dataset and incorporate semantic information into the edges of the graph structure to enable more fine-grained evaluation.

\textbf{Reproducibility and Asset Use.} The appendix provides construction statistics, prompts, sample questions, graph examples, robustness settings, and the evaluation environment. Anonymized benchmark data, graph construction code, evaluation scripts, and Croissant metadata are provided in the anonymized repository linked in the main text. The anonymized release includes Croissant metadata with license and Responsible AI fields. Existing models, APIs, and source literature are cited in the paper and are used under their original access terms.

\subsection{LLM Usage}
\label{app:llm_usage}

LLMs are used as part of the benchmark construction pipeline for document analysis, guided question generation, keyword extraction, and multi-model committee assessment, as described in Section~\ref{sec:graph-construction} and Appendix~\ref{ap1}. A large language model was also used for drafting and language polishing of this article. The authors are fully responsible for the accuracy, integrity, and interpretation of the work.

\subsection{Details of LG-Bench}\label{ap1}

\subsubsection{Hyperparameter Settings}
Each edge in our evaluation graph represents a weighted semantic relationship between questions, computed using our bidirectional matching and core similarity algorithms. For graph construction, we set the balance parameter $\gamma = 0.6$ to weight core similarity slightly higher than bidirectional matching, selected the top $\kappa = 3$ strongest connections for core similarity computation, and applied an adaptive edge threshold $\theta = 0.4$ to ensure meaningful semantic relationships while maintaining graph connectivity.

\subsubsection{Detailed Statistics of Dataset Construction}
\label{app:dataset_stats}

To ensure the high reliability of our benchmark, we implemented a strict multi-stage filtering process. While the methodology is detailed in the main text, this appendix provides the quantitative breakdown of the data cleaning and quality control phases.

\textbf{Attrition Analysis.} The construction pipeline followed a funnel structure, transitioning from raw generation to automated committee filtering, and finally to human expert validation. Table \ref{tab:construction_stats} details the volume of questions processed at each stage.

Initially, the system generated a raw corpus of 13,250 questions. The Multi-Model Committee acted as the first line of defense, filtering out 2,653 low-quality candidates (approximately 20\%) that exhibited structural flaws, shallowness, or a lack of clear textual evidence. This resulted in a candidate set of 10,597 questions.

In the final stage, human experts conducted a targeted verification process. Specifically, experts verified a total of 4,153 questions, which included candidates explicitly flagged by the Multi-Model Committee and a random sample of the remaining questions. We retained the annotation records during the review, and the calculated Fleiss' Kappa of 0.86 indicates a high level of agreement. This rigorous review led to the removal of an additional 397 questions. Consequently, the final curated dataset consists of 10,200 high-quality, validated questions.

\begin{table}[h]
    \centering
    \caption{\textbf{Statistics of the Data Construction Pipeline.} The table tracks the number of questions retained and removed at each stage of the quality control process.}
    \label{tab:construction_stats}
    \begin{tabular}{l l r r}
        \toprule
        \textbf{Stage} & \textbf{Description}            & \textbf{Count}  & \textbf{Removed} \\
        \midrule
        Stage 1        & Initially generated questions   & 13,250          & -                \\
        Stage 2        & Multi-Model Committee Selection & 10,597          & 2,653            \\
        Stage 3        & Expert Review                   & \textbf{10,200} & 397              \\
        \bottomrule
    \end{tabular}
\end{table}

\textbf{Error Distribution Analysis.} To better understand the limitations of the automated generation and committee phases, we analyzed the reasons for rejection during the final expert validation. As shown in Table \ref{tab:error_types}, the 397 rejected questions were categorized into three primary error types.

\begin{table}[h]
    \centering
    \caption{\textbf{Distribution of Error Types in Expert Validation.} Analysis of the 397 questions rejected during the final human review stage.}
    \label{tab:error_types}
    \begin{tabular}{l r r}
        \toprule
        \textbf{Error Type}                & \textbf{Number} & \textbf{Percentage} \\
        \midrule
        Low relevance to the domain        & 214             & 53.9\%              \\
        Topics not suitable for evaluation & 136             & 34.3\%              \\
        Logical or factual errors          & 47              & 11.8\%              \\
        \midrule
        \textbf{Total Rejected}            & \textbf{397}    & \textbf{100.0\%}    \\
        \bottomrule
    \end{tabular}
\end{table}

Notably, low relevance (214 cases) was the most frequent reason for rejection, referring to questions that were technically correct but focused on trivial details rather than the core scientific contributions of the paper. Logical or factual errors constituted the smallest category (47 cases), suggesting that the preceding Multi-Model Committee was highly effective in filtering out hallucinations and objective inaccuracies before human review.

\subsubsection{Prompt Design for LG-Bench Construction}

The prompt design for the LG-Bench construction pipeline follows a structured multi-agent approach. The process begins with the Document Analysis Agent, whose core system prompt is illustrated in Figure \ref{fig:agent1_part1}. This agent operates based on detailed difficulty instructions and Bloom's taxonomy levels, as further specified in Figure \ref{fig:agent1_part2}. The synthesized instructions and paper analysis are then passed to the Question Generator, which follows the system prompt and strict JSON output formatting requirements shown in Figure \ref{fig:agent2_prompt}. To ensure the quality and scientific accuracy of the generated items, each question is evaluated through a Multi-Model Committee Assessment, guided by the multi-dimensional evaluation criteria and scoring rubric presented in Figure \ref{fig:evaluation_prompt}.

\subsubsection{Sample Questions and Graph Structure from LG-Bench}

\textbf{Sample Questions.} The following section presents several representative examples from our dataset, including options and explanations. Correct answers are highlighted in red.

\noindent\rule{\linewidth}{0.4pt}  
\textbf{Question 48:} What primary cellular phenotype links loss of keratinization genes to increased EAC aggressiveness?

\begin{itemize}
    \item (A) Suppression of Wnt signaling curtailing stem-cell renewal
    \item \textcolor{red}{\textbf{(B) Enhanced epithelial plasticity facilitating invasion}}
    \item (C) Elevated oxidative respiration reducing hypoxia tolerance
    \item (D) Activation of DNA damage checkpoints halting proliferation
\end{itemize}

\textbf{Explanation:} Loss of terminal squamous differentiation increases plasticity, enabling invasion and aggressive tumor behavior.

\vspace{0.5em}
\noindent\rule{\linewidth}{0.4pt}


\textbf{Question 915:} If imaging-only maps recover large assemblies better than small ones, what physical explanation best accounts for this?

\begin{itemize}
    \item (A) Small complexes have lower protein copy numbers, depleting peptide coverage in MS.
    \item \textcolor{red}{\textbf{(B) Spatial resolution of light microscopy limits detection of nanometre-scale complexes, causing small assemblies to be visually indistinguishable.}}
    \item (C) Image segmentation algorithms preferentially crop large structures for analysis.
    \item (D) Large assemblies generate brighter fluorescence signal enabling better antibody binding.
\end{itemize}

\textbf{Explanation:} Optical diffraction restricts resolution (~200 nm); small complexes below this scale cannot be visually separated, whereas larger assemblies occupy resolvable regions.

\vspace{0.5em}
\noindent\rule{\linewidth}{0.4pt}

\textbf{Question 1767:} When donor-to-recipient cell ratios in MitoTRACER coculture increased from 1:10 to 1:1, the percentage of green-converted cancer cells rose. This demonstrates that mitochondrial transfer probability depends mainly on which factor?

\begin{itemize}
    \item \textcolor{red}{\textbf{(A) Donor cell abundance creating more contact opportunities}}
    \item (B) Recipient cell cell-cycle phase
    \item (C) Mitochondrial fission rate in recipients
    \item (D) Serum glucose concentration
\end{itemize}

\textbf{Explanation:} Higher donor proportion increases physical interactions and nanotube formation, elevating organelle transfer frequency.

\vspace{0.5em}
\noindent\rule{\linewidth}{0.4pt}

\textbf{Question 2148:} What experimental evidence argues that myddosome clearance is p62-dependent?

\begin{itemize}
    \item (A) p62 overexpression reduced IL-6 secretion
    \item (B) ProteoStat staining increased in p62-deficient cells
    \item \textcolor{red}{\textbf{(C) p62 knockout macrophages accumulated residual myddosome structures long after signaling subsided}}
    \item (D) Phospho-p65 levels declined faster in p62 knockout cells
\end{itemize}

\textbf{Explanation:} Persistence of remnants specifically in p62-null cells indicates its necessity for targeting complexes to autophagy.

\vspace{0.5em}
\noindent\rule{\linewidth}{0.4pt}

\textbf{Question 3992:} Stability studies demonstrated that folate-diketone remained bound to CovCAR at pH 4.5, whereas folate-FITC dissociated. This finding predicts which therapeutic advantage?

\begin{itemize}
    \item (A) Faster renal clearance of the adapter
    \item \textcolor{red}{\textbf{(B) Enhanced trafficking through acidic tumour microenvironments}}
    \item (C) Reduced requirement for lymphodepletion
    \item (D) Lower likelihood of cytokine release syndrome
\end{itemize}

\textbf{Explanation:} Retention of covalent linkage under acidic conditions supports sustained CAR-adapter association in endosomes and acidic tumour niches.

\vspace{0.5em}
\noindent\rule{\linewidth}{0.4pt}

\textbf{Question 4134:} Circular dichroism of the phenanthroline chromophore shows a negative Cotton effect at 330 nm for (+)-5-[Cu][Lu]. What stereochemical element is most directly inferred from this observation?

\begin{itemize}
    \item (A) $\Delta$-helicity at the Lu(III) helicate
    \item \textcolor{red}{\textbf{(B) $\Lambda$-helicity at the Cu(I)-dpp clasp}}
    \item (C) Presence of racemic knot mixture
    \item (D) Metal-free macrocycle formation
\end{itemize}

\textbf{Explanation:} A negative Cotton effect for bis-dpp Cu(I) complexes corresponds to $\Lambda$ helicity, indicating that the clasp crossing in the knot adopts $\Lambda$ configuration.

\vspace{0.5em}
\noindent\rule{\linewidth}{0.4pt}

\textbf{Question 7441:} To evaluate whether N-0385 can block influenza A virus entry in vitro, which cell model would best replicate the TMPRSS2-dependent activation step targeted, and why?

\begin{itemize}
    \item \textcolor{red}{\textbf{(A) Human Calu-3 airway epithelial cells, because they endogenously express TMPRSS2 and other TTSPs needed for viral fusion}}
    \item (B) BHK-21 hamster fibroblasts, because they lack serine proteases that interfere with fusion assays
    \item (C) Vero E6 kidney cells, because they are routinely used for high-titre virus propagation despite minimal TMPRSS2 expression
    \item (D) HEK-293 human embryonic kidney cells, because they overexpress ACE2 after transient transfection
\end{itemize}

\textbf{Explanation:} Calu-3 cells were used for their endogenous TMPRSS2 activity, allowing peptidomimetic inhibitors to block spike/hemagglutinin activation.

\vspace{0.5em}
\noindent\rule{\linewidth}{0.4pt}

\textbf{Question 8622:} Vorinostat targets class I/II HDACs. Which downstream effect most directly increases tumor immunogenicity?

\begin{itemize}
    \item (A) Blocking VEGF secretion, thereby reducing angiogenesis.
    \item \textcolor{red}{\textbf{(B) Enhanced histone acetylation leading to up-regulation of MHC class I genes.}}
    \item (C) Suppression of DNA repair enzymes, causing mutational overload.
    \item (D) Direct phosphorylation of STAT3, activating immune checkpoints.
\end{itemize}

\textbf{Explanation:} HDAC inhibition acetylates chromatin and increases expression of antigen-presentation molecules, improving immune recognition.

\vspace{0.5em}
\noindent\rule{\linewidth}{0.4pt}

\textbf{Question 9437:} Which control would MOST convincingly demonstrate that the loss of stress-granule assembly after RIOK1 shRNA is on-target?

\begin{itemize}
    \item (A) Overexpress GFP to control for lentiviral transduction.
    \item \textcolor{red}{\textbf{(B) Re-express an shRNA-resistant RIOK1 cDNA and test whether SG formation is rescued.}}
    \item (C) Add cycloheximide to dissolve granules in all samples.
    \item (D) Include a non-targeting shRNA vector in parallel cultures.
\end{itemize}

\textbf{Explanation:} Functional rescue with an shRNA-proof construct specifically attributes the phenotype to RIOK1 depletion.

\vspace{0.5em}
\noindent\rule{\linewidth}{0.4pt}

\textbf{Question 9885:} A researcher cultures iPSC-derived cardiac fibroblasts on 2 kPa hydrogels and treats them with exogenous TGF$\beta$. What combined manipulation would most effectively restore quiescence?

\begin{itemize}
    \item (A) Blockade of IL-1$\beta$ signalling with anakinra alone
    \item (B) Knockdown of YAP together with ROCK inhibition
    \item (C) Overexpression of SORBS2 together with blebbistatin
    \item \textcolor{red}{\textbf{(D) Addition of SB431542 together with saracatinib}}
\end{itemize}

\textbf{Explanation:} Soft substrate plus TGF$\beta$ inhibition alone was insufficient after activation; adding SRC inhibition (saracatinib) created the synergy required for full reversal toward quiescence.

\vspace{0.5em}
\noindent\rule{\linewidth}{0.4pt}

\begin{figure}[htbp]
    \centering
    \includegraphics[width=0.95\linewidth]{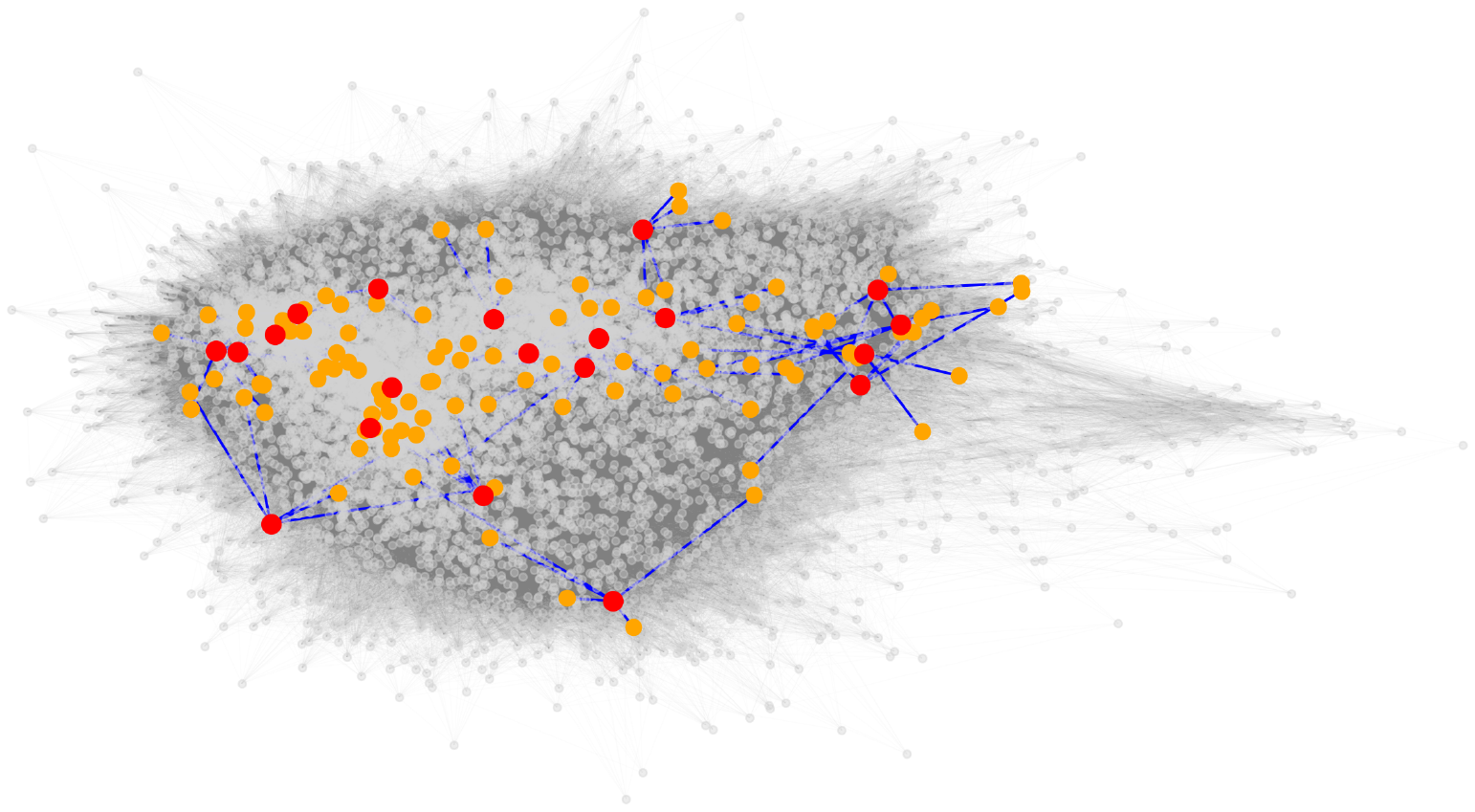}
    \caption{Visualization of a subset of LG-Bench, where red nodes represent sampled questions and their top-5 semantic neighbors (orange) based on edge weights. All other nodes and edges are displayed in gray to show overall connectivity. The blue edges indicate the strongest semantic connections (top-5 weighted edges) for each sampled node.}
    \label{fig:haha}
\end{figure}

\textbf{Sample Graph Structure.} We sample several local subgraph structures and visualize them, as shown in Fig.~\ref{fig:haha}. We alse present the neighboring information, as shown in Table~\ref{tab:haha}. It can be observed that the graph structure of our dataset captures rich semantic connections.

\begin{table}[t]
    \centering
    \caption{Top-5 highest-weight neighbors of 20 randomly sampled nodes. Weight (w) indicates semantic similarity.}
    \resizebox{\linewidth}{!}{
        \begin{tabular}{|c|c|c|c|c|c|c|}
            \hline
            \textbf{Node ID} & \textbf{Degree} & \textbf{Neighbor 1 (w)} & \textbf{Neighbor 2 (w)} & \textbf{Neighbor 3 (w)} & \textbf{Neighbor 4 (w)} & \textbf{Neighbor 5 (w)} \\
            \hline
            48               & 191             & 39 (0.6531)             & 465 (0.6498)            & 9952 (0.6269)           & 497 (0.6089)            & 29 (0.5966)             \\
            4552             & 239             & 4564 (0.6253)           & 4538 (0.6165)           & 4524 (0.6103)           & 4587 (0.5998)           & 8973 (0.5913)           \\
            7441             & 204             & 7448 (0.8288)           & 7690 (0.7892)           & 2851 (0.7648)           & 7685 (0.7440)           & 7445 (0.7353)           \\
            2822             & 471             & 2824 (0.8452)           & 2828 (0.8393)           & 3419 (0.8200)           & 2829 (0.7956)           & 3416 (0.7751)           \\
            915              & 478             & 2446 (0.7431)           & 10071 (0.7266)          & 6640 (0.6843)           & 8843 (0.6815)           & 3149 (0.6744)           \\
            9216             & 61              & 9086 (0.6868)           & 9078 (0.6770)           & 9061 (0.6594)           & 9101 (0.6590)           & 9211 (0.6177)           \\
            9885             & 100             & 9735 (0.7791)           & 9883 (0.7493)           & 9881 (0.7485)           & 9732 (0.7007)           & 9739 (0.6861)           \\
            9437             & 195             & 9032 (0.7917)           & 9045 (0.7405)           & 9022 (0.7328)           & 7376 (0.7022)           & 9012 (0.6970)           \\
            9604             & 297             & 9610 (0.8237)           & 9955 (0.8006)           & 9606 (0.7000)           & 7740 (0.6839)           & 7753 (0.6775)           \\
            4134             & 66              & 4138 (0.6928)           & 4171 (0.6872)           & 4168 (0.6585)           & 4180 (0.6016)           & 4144 (0.5986)           \\
            2707             & 113             & 5536 (0.7635)           & 3346 (0.7352)           & 5531 (0.6819)           & 3341 (0.6663)           & 5899 (0.6505)           \\
            388              & 187             & 528 (0.8545)            & 513 (0.8285)            & 379 (0.8248)            & 550 (0.7506)            & 548 (0.7412)            \\
            2729             & 104             & 2752 (0.8181)           & 3320 (0.7863)           & 2739 (0.7591)           & 2714 (0.7580)           & 2721 (0.7579)           \\
            8047             & 116             & 8574 (0.7978)           & 8044 (0.6790)           & 8595 (0.6327)           & 8588 (0.6265)           & 8058 (0.6261)           \\
            1767             & 484             & 2084 (0.8085)           & 2066 (0.8027)           & 2064 (0.7842)           & 1785 (0.7794)           & 1795 (0.7515)           \\
            8622             & 598             & 8616 (0.8116)           & 8062 (0.7839)           & 8086 (0.7676)           & 8614 (0.7465)           & 8633 (0.7454)           \\
            9528             & 314             & 7740 (0.8151)           & 7734 (0.7978)           & 9410 (0.7548)           & 7739 (0.6999)           & 8164 (0.6966)           \\
            758              & 84              & 784 (0.7270)            & 753 (0.6624)            & 776 (0.6561)            & 783 (0.6472)            & 760 (0.6453)            \\
            2148             & 587             & 2108 (0.8436)           & 2016 (0.8381)           & 2001 (0.8345)           & 2131 (0.8325)           & 2129 (0.8256)           \\
            3992             & 169             & 3983 (0.7378)           & 3996 (0.7156)           & 807 (0.6854)            & 4007 (0.6290)           & 4005 (0.6283)           \\
            \hline
        \end{tabular}
    }
    \label{tab:haha}
\end{table}

\subsection{Details of Evaluation}\label{ap2}

\subsubsection{Evaluation Environment}
\label{app:evaluation_environment}
For models with open weights, we downloaded the official weights and deployed tests on a cluster with 8 nodes as shown in the Table~\ref{tab:expenv}. We set the same default temperature parameters and tested with the exact same prompt. For models with closed weights, we conducted tests based on the API while maintaining the same testing configurations as the open-weight models to ensure fairness.

\begin{table}[!h]
    \centering
    \caption{Experimental Environment}
    \begin{tabular}{@{}ll@{}}
        \toprule
        \textbf{Component}  & \textbf{Specification}          \\
        \midrule

        \multicolumn{2}{@{}l@{}}{\textbf{CPU}}                \\
        \midrule

        Model               & Intel(R) Xeon(R) Platinum 8336C \\
        Total Cores         & 128                             \\
        Total Threads       & 128                             \\
        Max Turbo Frequency & 2.30GHz                         \\
        \midrule

        \multicolumn{2}{@{}l@{}}{\textbf{GPU}}                \\
        \midrule

        Model               & NVIDIA A800-SXM4 $\times 8$     \\
        VRAM                & 80 GB GDDR6X                    \\
        \bottomrule
    \end{tabular}

    \label{tab:expenv}
\end{table}

\subsubsection{Robustness Experiment Details}
\label{Robustness}

To provide a comprehensive view of our sensitivity analysis, we present the complete experimental results for varying graph sparsity levels ($\theta$) and connectivity weights ($\gamma$). Tables~\ref{tab:appendix_theta05_full} and \ref{tab:appendix_theta06_full} detail the performance metrics for $\theta=0.5$ and $\theta=0.6$, respectively. Consistent with the main text, we separate the results by the connectivity parameter $\gamma$ to isolate the effects of bidirectional versus core similarity weighting.

Table~\ref{tab:appendix_theta05_full} illustrates the medium sparsity setting ($\theta=0.5$), comparing $\gamma=0.4$ (Average Degree $\approx 110.75$) and $\gamma=0.6$ (Average Degree $\approx 119.17$). Table~\ref{tab:appendix_theta06_full} presents the high sparsity setting ($\theta=0.6$), comparing $\gamma=0.4$ (Average Degree $\approx 23.98$) and $\gamma=0.6$ (Average Degree $\approx 24.12$). In all configurations, while the absolute values of KBS shift due to changes in graph density, the relative performance hierarchy of the models remains stable, confirming the robustness of the LG-Bench diagnostic framework.

\begin{table}[h!]
    \centering
    \caption{\textbf{Detailed Robustness Results at Medium Sparsity ($\theta = 0.5$).} We compare graph connectivity settings with a stricter similarity threshold than the default. Table (a) uses $\gamma = 0.4$, and Table (b) uses $\gamma = 0.6$. The omission of the accuracy metric allows for a focused analysis of graph-based coherence (GCS) and balance (KBS).}
    \label{tab:appendix_theta05_full}

    \textbf{(a) Setting $\theta = 0.5, \gamma = 0.4$ (Avg. Degree $\approx$ 110.75)} \\
    \resizebox{\linewidth}{!}{
        \begin{tabular}{l|cc|cc|cc|cc}
            \toprule
            \multirow{2}{*}{\textbf{Model}} & \multicolumn{2}{c|}{\textbf{Medicine}} & \multicolumn{2}{c|}{\textbf{Biology}} & \multicolumn{2}{c|}{\textbf{Chemistry}} & \multicolumn{2}{c}{\textbf{Overall}}                                                             \\
            \cmidrule(lr){2-3} \cmidrule(lr){4-5} \cmidrule(lr){6-7} \cmidrule(lr){8-9}
                                            & \textbf{GCS}                           & \textbf{KBS}                          & \textbf{GCS}                            & \textbf{KBS}                         & \textbf{GCS} & \textbf{KBS} & \textbf{GCS} & \textbf{KBS} \\
            \midrule
            Qwen2.5-7B                      & 62.63                                  & 124.92                                & 64.11                                   & 77.40                                & 59.29        & 89.46        & 62.23        & 64.70        \\
            Llama-3-8B                      & 54.40                                  & 142.21                                & 57.50                                   & 61.07                                & 53.16        & 99.42        & 55.32        & 63.26        \\
            Llama3-Med42-8B                 & 61.78                                  & 117.57                                & 60.14                                   & 63.90                                & 56.87        & 104.30       & 59.64        & 58.67        \\
            Qwen2.5-14B                     & 69.64                                  & 95.62                                 & 72.70                                   & 60.96                                & 68.92        & 71.85        & 70.60        & 44.11        \\
            Qwen2.5-32B                     & 72.59                                  & 101.22                                & 74.37                                   & 57.84                                & 70.63        & 71.71        & 72.81        & 45.99        \\
            Llama-3.1-70B                   & 69.62                                  & 106.73                                & 69.20                                   & 77.25                                & 67.01        & 83.49        & 68.81        & 53.04        \\
            Qwen2.5-72B                     & 75.14                                  & 89.02                                 & 75.71                                   & 57.89                                & 72.77        & 77.55        & 74.78        & 43.04        \\
            \bottomrule
        \end{tabular}
    }

    \vspace{1.5em}

    \textbf{(b) Setting $\theta = 0.5, \gamma = 0.6$ (Avg. Degree $\approx$ 119.17)} \\
    \resizebox{\linewidth}{!}{
        \begin{tabular}{l|cc|cc|cc|cc}
            \toprule
            \multirow{2}{*}{\textbf{Model}} & \multicolumn{2}{c|}{\textbf{Medicine}} & \multicolumn{2}{c|}{\textbf{Biology}} & \multicolumn{2}{c|}{\textbf{Chemistry}} & \multicolumn{2}{c}{\textbf{Overall}}                                                             \\
            \cmidrule(lr){2-3} \cmidrule(lr){4-5} \cmidrule(lr){6-7} \cmidrule(lr){8-9}
                                            & \textbf{GCS}                           & \textbf{KBS}                          & \textbf{GCS}                            & \textbf{KBS}                         & \textbf{GCS} & \textbf{KBS} & \textbf{GCS} & \textbf{KBS} \\
            \midrule
            Qwen2.5-7B                      & 63.12                                  & 127.43                                & 64.58                                   & 79.62                                & 59.78        & 91.77        & 62.71        & 66.21        \\
            Llama-3-8B                      & 54.89                                  & 145.09                                & 57.96                                   & 63.29                                & 53.67        & 101.82       & 55.83        & 65.35        \\
            Llama3-Med42-8B                 & 62.28                                  & 119.93                                & 60.62                                   & 66.27                                & 57.38        & 106.81       & 60.15        & 60.47        \\
            Qwen2.5-14B                     & 70.14                                  & 97.81                                 & 73.18                                   & 63.11                                & 69.44        & 74.07        & 71.11        & 46.00        \\
            Qwen2.5-32B                     & 73.09                                  & 103.50                                & 74.87                                   & 59.92                                & 71.15        & 71.70        & 73.33        & 47.81        \\
            Llama-3.1-70B                   & 70.11                                  & 109.12                                & 69.71                                   & 79.43                                & 67.54        & 85.32        & 69.34        & 55.41        \\
            Qwen2.5-72B                     & 75.64                                  & 91.17                                 & 76.21                                   & 60.03                                & 73.30        & 79.23        & 75.31        & 45.21        \\
            \bottomrule
        \end{tabular}
    }
\end{table}

\begin{table}[h!]
    \centering
    \caption{\textbf{Detailed Robustness Results at High Sparsity ($\theta = 0.6$).} We examine metric stability under highly restrictive graph structures. Table (c) uses $\gamma = 0.4$, and Table (d) uses $\gamma = 0.6$.}
    \label{tab:appendix_theta06_full}

    \textbf{(c) Setting $\theta = 0.6, \gamma = 0.4$ (Avg. Degree $\approx$ 23.98)} \\
    \resizebox{\linewidth}{!}{
        \begin{tabular}{l|cc|cc|cc|cc}
            \toprule
            \multirow{2}{*}{\textbf{Model}} & \multicolumn{2}{c|}{\textbf{Medicine}} & \multicolumn{2}{c|}{\textbf{Biology}} & \multicolumn{2}{c|}{\textbf{Chemistry}} & \multicolumn{2}{c}{\textbf{Overall}}                                                             \\
            \cmidrule(lr){2-3} \cmidrule(lr){4-5} \cmidrule(lr){6-7} \cmidrule(lr){8-9}
                                            & \textbf{GCS}                           & \textbf{KBS}                          & \textbf{GCS}                            & \textbf{KBS}                         & \textbf{GCS} & \textbf{KBS} & \textbf{GCS} & \textbf{KBS} \\
            \midrule
            Qwen2.5-7B                      & 61.94                                  & 262.53                                & 63.79                                   & 155.60                               & 58.74        & 215.85       & 62.09        & 134.09       \\
            Llama-3-8B                      & 53.61                                  & 301.92                                & 57.15                                   & 125.30                               & 52.81        & 242.89       & 55.14        & 139.71       \\
            Llama3-Med42-8B                 & 60.97                                  & 259.10                                & 59.88                                   & 127.72                               & 56.51        & 272.32       & 59.51        & 129.37       \\
            Qwen2.5-14B                     & 68.98                                  & 218.37                                & 72.40                                   & 129.84                               & 68.34        & 179.09       & 70.47        & 95.29        \\
            Qwen2.5-32B                     & 72.01                                  & 219.29                                & 74.18                                   & 115.46                               & 70.21        & 168.08       & 72.76        & 95.15        \\
            Llama-3.1-70B                   & 68.95                                  & 228.51                                & 68.95                                   & 155.13                               & 66.78        & 198.68       & 68.82        & 111.08       \\
            Qwen2.5-72B                     & 74.32                                  & 201.61                                & 75.40                                   & 119.89                               & 72.36        & 180.75       & 74.65        & 91.82        \\
            \bottomrule
        \end{tabular}
    }

    \vspace{1.5em}

    \textbf{(d) Setting $\theta = 0.6, \gamma = 0.6$ (Avg. Degree $\approx$ 24.12)} \\
    \resizebox{\linewidth}{!}{
        \begin{tabular}{l|cc|cc|cc|cc}
            \toprule
            \multirow{2}{*}{\textbf{Model}} & \multicolumn{2}{c|}{\textbf{Medicine}} & \multicolumn{2}{c|}{\textbf{Biology}} & \multicolumn{2}{c|}{\textbf{Chemistry}} & \multicolumn{2}{c}{\textbf{Overall}}                                                             \\
            \cmidrule(lr){2-3} \cmidrule(lr){4-5} \cmidrule(lr){6-7} \cmidrule(lr){8-9}
                                            & \textbf{GCS}                           & \textbf{KBS}                          & \textbf{GCS}                            & \textbf{KBS}                         & \textbf{GCS} & \textbf{KBS} & \textbf{GCS} & \textbf{KBS} \\
            \midrule
            Qwen2.5-7B                      & 62.48                                  & 261.28                                & 64.21                                   & 156.73                               & 58.95        & 217.12       & 62.84        & 135.38       \\
            Llama-3-8B                      & 53.28                                  & 300.45                                & 56.89                                   & 126.58                               & 53.47        & 241.73       & 55.92        & 140.86       \\
            Llama3-Med42-8B                 & 61.73                                  & 258.15                                & 59.56                                   & 128.91                               & 57.12        & 271.48       & 60.35        & 130.62       \\
            Qwen2.5-14B                     & 69.65                                  & 217.24                                & 72.78                                   & 131.15                               & 68.92        & 180.38       & 71.23        & 96.78        \\
            Qwen2.5-32B                     & 71.58                                  & 220.46                                & 74.83                                   & 116.72                               & 70.94        & 169.54       & 73.42        & 96.19        \\
            Llama-3.1-70B                   & 68.47                                  & 227.38                                & 69.71                                   & 156.47                               & 67.35        & 199.82       & 69.58        & 112.35       \\
            Qwen2.5-72B                     & 74.86                                  & 200.73                                & 75.92                                   & 121.24                               & 72.98        & 181.93       & 75.38        & 93.15        \\
            \bottomrule
        \end{tabular}
    }
\end{table}

\begin{figure}[t]
    \centering
    \begin{tcolorbox}[
            colback=orange!3!white,
            colframe=orange!60!black,
            title=\textbf{Prompt for Document Analysis Agent},
            fonttitle=\bfseries,
            boxrule=0.8pt,
            arc=3pt,
            left=6pt,
            right=6pt,
            top=6pt,
            bottom=6pt
        ]
        \scriptsize 
        \texttt{You are an expert content creator specializing in designing domain-specific, knowledge-based assessments for life sciences.}

        \vspace{4pt}
        \texttt{You will be provided with:}
        \begin{enumerate}[leftmargin=1.5em, itemsep=0pt, topsep=2pt]
            \item \texttt{A research paper in the life sciences (e.g., biology, medicine, pharmacology)}
            \item \texttt{A target difficulty level: \{difficulty\_level\}}
            \item \texttt{Bloom's taxonomy levels: \{bloom\_levels\}}
            \item \texttt{Difficulty description: \{difficulty\_description\}}
        \end{enumerate}

        \vspace{4pt}
        \texttt{\textbf{Task:}}
        \texttt{Your task is to generate a precise, instructional prompt that will guide another AI agent to create exactly \{num\_questions\} high-quality multiple-choice questions (MCQs) at the specified \{difficulty\_level\}. The resulting questions must assess biologically meaningful knowledge and scientific understanding derived from the paper.}

        \vspace{4pt}
        \texttt{\textbf{Important Global Rules:}}
        \begin{enumerate}[leftmargin=1.5em, itemsep=3pt, topsep=2pt]
            \item \texttt{\textbf{Stand-Alone Questions:} Question stems must read as self-contained science items.}
                  \begin{itemize}[leftmargin=1em, label={$\bullet$}]
                      \item \texttt{DO NOT use phrases like "in the paper," "according to the study," "as described by the authors," etc.}
                      \item \texttt{Embed the scientific knowledge naturally, so questions look like a conventional test bank rather than a reading-comprehension quiz.}
                  \end{itemize}

            \item \texttt{\textbf{Focus Scope:}}
                  \begin{itemize}[leftmargin=1em, label={$\bullet$}]
                      \item \texttt{Biological, biochemical, medical, or pharmacological concepts introduced or explained in the paper.}
                      \item \texttt{Avoid trivialities (dataset names, figure numbers, author affiliations).}
                  \end{itemize}

            \item \texttt{\textbf{Question Types \& Cognitive Skills:}}
                  \begin{itemize}[leftmargin=1em, label={$\bullet$}]
                      \item \texttt{Knowledge-oriented: definitions, classifications, functions, relationships, mechanisms.}
                      \item \texttt{Align with \{bloom\_levels\} for the given \{difficulty\_level\}.}
                      \item \texttt{Test recognition, comprehension, or application (depending on difficulty), not rote details.}
                  \end{itemize}

            \item \texttt{\textbf{Distractor Quality:}}
                  \begin{itemize}[leftmargin=1em, label={$\bullet$}]
                      \item \texttt{Scientifically plausible, relevant, and non-obvious.}
                      \item \texttt{Leverage common misconceptions or closely related concepts to raise challenge.}
                  \end{itemize}

            \item \texttt{\textbf{Suggested Concept Areas to Mine from the Paper:}}
                  \begin{itemize}[leftmargin=1em, label={$\bullet$}]
                      \item \texttt{Functional roles of genes/proteins/pathways.}
                      \item \texttt{Mechanisms of disease or therapeutic action.}
                      \item \texttt{Diagnostic markers, disease progression indicators.}
                      \item \texttt{Significant experimental findings establishing cause--effect or structure--function relationships.}
                      \item \texttt{Key concepts from molecular biology, neuroscience, pharmacology, genetics, etc.}
                  \end{itemize}

            \item \texttt{\textbf{Output Requirements:} Do not include or request generation templates or examples.}
        \end{enumerate}

        \vspace{4pt}
        \texttt{\textbf{Paper Analysis Guidelines (for the question-generating agent):}}
        \begin{itemize}[leftmargin=1.5em, itemsep=1pt, topsep=2pt]
            \item \texttt{Extract core definitions, mechanisms, and terminologies essential for understanding.}
            \item \texttt{Identify clear cause--effect, structure--function, or mechanistic explanations suitable for MCQs.}
            \item \texttt{Skip overly technical math or procedural minutiae with low educational value.}
        \end{itemize}

        \vspace{4pt}
        \texttt{\textbf{Final Deliverable (for the question-generating agent):}}
        \texttt{Produce precisely \{num\_questions\} MCQs that comply with all rules above and match the specified \{difficulty\_level\} and \{bloom\_levels\}. No additional commentary or formatting instructions are needed here; follow your own output template.}
    \end{tcolorbox}
    \caption{Prompt for Document Analysis Agent.}
    \label{fig:agent1_part1}
\end{figure}

\subsubsection{Case Study}
To further analyse the performance differences between pre-trained language models of varying sizes in complex question-answering tasks, we selected a set of highly relevant questions as evaluation samples to compare the response quality of large models (Qwen-72B) and medium-sized models (7B) under the same input settings. We retained the original question stems, concealed the multiple-choice options, and prompted the models to perform open-ended question answering. Qwen-72B can answer this set of questions, consistently generating responses with clear structural logic, accurate content alignment, and a deep understanding of domain-specific mechanisms. In contrast, while the 7B model's responses demonstrate some structural and organisational coherence, they still exhibit errors in certain questions, with significantly insufficient depth of understanding, expression precision, and knowledge mobilisation capabilities. For example, in
the id2016 question, Qwen-7B overlooks the core mechanism of ``ubiquitin-dependent transfer to autophagy'' and instead provides a vague ``autophagy-related'' explanation. Additionally, the 7B model disregarded the role of the ``MyD88-dependent myddosome complex'' and erroneously assumed it influences early ``NF-$\kappa$B-dependent transcription''.
\\This misunderstanding recurs in similar questions or adjacent nodes, indicating the model struggles to effectively reuse contextual information and lacks a robust representation of the internal structure of pathway mechanisms. Furthermore, larger-scale models demonstrate stronger performance in contextual understanding and domain knowledge integration, while smaller-scale models often misidentify key mechanisms. This may stem from their insufficient understanding of domain-specific signalling pathways and their inability to distinguish between mechanistically similar components.
Despite the five samples sharing high-frequency domain-specific keywords—such as ``MyD88'', ``traf6'', and ``nf-$\kappa$B'',the smaller models (Qwen2.5-7B) still cannot consistently generalise mechanisms across samples. This inconsistency manifests not only as factual errors but also as failures in mechanism inference-for instance, missing prior knowledge of the``ubiquitin-mediated autophagy pathway'' in related cases  (id2032 and id2035). As the number of parameters expanded from 7B to 72B, the Qwen model showed significant improvements in consistency and depth in its understanding of professional mechanisms. This finding aligns with the conclusions drawn in the main text regarding differences in knowledge organization coherence, thereby substantiating the efficacy of graph structure indicators in differentiating model capabilities.


\begin{figure}[t]
    \centering
    \begin{tcolorbox}[
            colback=orange!3!white,
            colframe=orange!60!black,
            title=\textbf{Detailed difficulty instructions provided to Document Analysis Agent},
            fonttitle=\bfseries,
            boxrule=0.8pt,
            arc=3pt,
            left=6pt,
            right=6pt,
            top=6pt,
            bottom=6pt
        ]

        \texttt{The following difficulty descriptions are injected into the Agent 1 prompt based on the user's configuration.}

        \vspace{6pt}
        \texttt{\textbf{Difficulty Instructions: EASY}}
        \begin{itemize}[leftmargin=1.5em, itemsep=2pt, topsep=2pt]
            \item \texttt{Focus on basic recall of key terms, definitions, and fundamental concepts}
            \item \texttt{Test understanding of main ideas and straightforward interpretations}
            \item \texttt{Questions should assess recognition and comprehension of explicitly stated information}
            \item \texttt{Use simple, direct question stems like "What is...", "Which of the following defines...", "Which molecule..."}
            \item \texttt{Distractors should be obviously incorrect to someone who read the paper carefully}
            \item \texttt{Avoid complex analysis or evaluation requirements}
        \end{itemize}

        \vspace{6pt}
        \hrule
        \vspace{6pt}

        \texttt{\textbf{Difficulty Instructions: MEDIUM}}
        \begin{itemize}[leftmargin=1.5em, itemsep=2pt, topsep=2pt]
            \item \texttt{Focus on application of concepts to new scenarios and analysis of relationships}
            \item \texttt{Test ability to interpret data, compare approaches, and analyze methodologies}
            \item \texttt{Questions should require connecting ideas and applying knowledge to solve problems}
            \item \texttt{Use question stems like "How would...", "What would happen if...", "Compare...", "Analyze..."}
            \item \texttt{Distractors should be plausible but require deeper understanding to eliminate}
            \item \texttt{Include questions about methodology application and result interpretation}
        \end{itemize}

        \vspace{6pt}
        \hrule
        \vspace{6pt}

        \texttt{\textbf{Difficulty Instructions: HARD}}
        \begin{itemize}[leftmargin=1.5em, itemsep=2pt, topsep=2pt]
            \item \texttt{Focus on critical evaluation, synthesis, and creation of new understanding}
            \item \texttt{Test ability to critique approaches, evaluate validity, and propose alternatives}
            \item \texttt{Questions should require high-level reasoning and complex cognitive processing}
            \item \texttt{Use question stems like "Evaluate...", "Critique...", "Design...", "What are the implications..."}
            \item \texttt{Distractors should be sophisticated and require expert-level knowledge to eliminate}
            \item \texttt{Include questions about limitations, future directions, and theoretical implications}
        \end{itemize}

    \end{tcolorbox}
    \caption{Detailed difficulty instructions provided to Document Analysis Agent.}
    \label{fig:agent1_part2}
\end{figure}

\begin{figure}[t]
    \centering
    \begin{tcolorbox}[
            colback=purple!3!white,
            colframe=purple!60!black,
            title=\textbf{Prompt for Question Generator},
            fonttitle=\bfseries,
            boxrule=0.8pt,
            arc=3pt,
            left=6pt,
            right=6pt,
            top=6pt,
            bottom=6pt
        ]
        \scriptsize
        \texttt{You are an expert educational assessment creator with deep knowledge in academic research and pedagogy. You have been provided with specific instructions for creating questions at a particular difficulty level. Follow these instructions precisely.}

        \vspace{4pt}
        \texttt{\textbf{CRITICAL:} Your response must be valid JSON format following the exact structure specified below.}

        \vspace{4pt}
        \texttt{\textbf{Requirements for each question:}}
        \begin{enumerate}[leftmargin=1.5em, itemsep=2pt, topsep=4pt]
            \item \texttt{Question stem: Clear, focused, and directly related to paper content.}
            \item \texttt{Four answer choices: One correct answer and three plausible distractors.}
            \item \texttt{Appropriate difficulty: Match the specified cognitive level exactly.}
            \item \texttt{Accuracy: Ensure all information is correct and verifiable from the paper.}
            \item \texttt{Clarity: Use precise academic language appropriate for the field.}
        \end{enumerate}

        \vspace{4pt}
        \texttt{\textbf{JSON Output Format (MANDATORY):}}
        \begin{verbatim}
{
  "questions": [
    {
      "id": 1,
      "question": "Question text here",
      "options": {
        "A": "Option A text",
        "B": "Option B text",
        "C": "Option C text",
        "D": "Option D text"
      },
      "correct_answer": "X(One of A, B, C, D)",
      "explanation": "Brief explanation of why this answer is correct",
      "bloom_level": 2
    }
  ]
}
\end{verbatim}

        \vspace{4pt}
        \texttt{\textbf{Important JSON Requirements:}}
        \begin{itemize}[leftmargin=1.5em, itemsep=2pt, topsep=4pt]
            \item \texttt{Generate exactly \{num\_questions\} questions in the "questions" array.}
            \item \texttt{Each question must have a unique "id" from 1 to \{num\_questions\}.}
            \item \texttt{"correct\_answer" must be one of: "A", "B", "C", "D".}
            \item \texttt{"bloom\_level" must be an integer 1-6.}
            \item \texttt{Ensure valid JSON syntax with proper quotes, commas, and brackets.}
            \item \texttt{Do not include any text outside the JSON structure.}
        \end{itemize}

        \vspace{4pt}
        \texttt{Generate exactly \{num\_questions\} multiple-choice questions following the provided difficulty-specific instructions in valid JSON format.}
    \end{tcolorbox}

    \caption{Prompt for Question Generator.}
    \label{fig:agent2_prompt}
\end{figure}

\begin{figure}[t]
    \centering
    \begin{tcolorbox}[
            colback=blue!3!white,
            colframe=blue!60!black,
            title=\textbf{Prompt for Multi-Model Committee Assessment},
            fonttitle=\bfseries,
            boxrule=0.8pt,
            arc=3pt,
            left=6pt,
            right=6pt,
            top=6pt,
            bottom=6pt
        ]
        \scriptsize
        \texttt{You are an expert educational assessment evaluator specializing in scientific multiple-choice questions.}

        \vspace{4pt}
        \texttt{Please evaluate the following multiple-choice question across four dimensions and provide scores from 0.0 to 1.0 for each:}

        \vspace{4pt}
        \texttt{\textbf{QUESTION TO EVALUATE:}} \\
        \texttt{Question: \{question\}} \\
        \texttt{Options: A) \{option\_A\} B) \{option\_B\} C) \{option\_C\} D) \{option\_D\}} \\
        \texttt{Correct Answer: \{correct\_answer\} | Explanation: \{explanation\} | Bloom Level: \{bloom\_level\}}

        \vspace{4pt}
        \texttt{\textbf{EVALUATION CRITERIA:}}
        \begin{enumerate}[leftmargin=1.5em, itemsep=2pt, topsep=2pt]
            \item \texttt{\textbf{Question Quality (0.0-1.0)}: Is the question clear, unambiguous, and well-structured? Does it match the stated Bloom's taxonomy level? Is the difficulty appropriate and consistent?}
            \item \texttt{\textbf{Option Quality (0.0-1.0)}: Are the distractors plausible and scientifically reasonable? Is the correct answer clearly the best choice? Are options of similar length and complexity?}
            \item \texttt{\textbf{Paper Alignment (0.0-1.0)}: Does the question accurately reflect content from the provided research paper? Are the concepts and facts mentioned in the question verifiable from the paper content? Does it test meaningful scientific knowledge rather than trivial details (author names, dataset names)? If paper content is provided, cross-reference the question against the actual paper text.}
            \item \texttt{\textbf{Assessment Suitability (0.0-1.0)}: Is this question appropriate for use in an evaluation dataset targeting model comprehension? Does it meaningfully discriminate between different levels of understanding or reasoning ability? Is the question well-formed and unambiguous, with minimal room for interpretation or dispute? If the question is based on specific content from a paper, is it fair in expecting the answer to be grounded in that content, without requiring outside knowledge or guesswork?}
        \end{enumerate}

        \vspace{4pt}
        \texttt{\textbf{OUTPUT FORMAT (REQUIRED):}}
        \begin{verbatim}
{
  "question_quality": 0.X,
  "option_quality": 0.X,
  "paper_alignment": 0.X,
  "assessment_suitability": 0.X,
  "overall_score": 0.X,
  "reasoning": "Brief explanation of the evaluation focusing on key strengths and weaknesses."
}
\end{verbatim}

        \vspace{4pt}
        \texttt{\textbf{IMPORTANT:}}
        \begin{itemize}[leftmargin=1.5em, itemsep=1pt, topsep=2pt]
            \item \texttt{Provide scores as decimals between 0.0 and 1.0.}
            \item \texttt{Overall score should be a weighted average based on the importance of each criterion.}
            \item \texttt{Keep reasoning concise but specific. Be critical but fair in your evaluation.}
            \item \texttt{If research paper content is provided above, use it to verify the accuracy and relevance of the question. For Paper Alignment scoring, cross-check facts and concepts against the actual paper content.}
        \end{itemize}

        \vspace{4pt}
        \texttt{\{paper\_content\}}
    \end{tcolorbox}
    \caption{Prompt for Multi-Model Committee Assessment.}
    \label{fig:evaluation_prompt}
\end{figure}

\end{document}